\documentclass[10pt,twocolumn,letterpaper]{article}

\usepackage[pagenumbers]{cvpr} 

\usepackage{graphicx}
\usepackage{amsmath}
\usepackage{amssymb}
\usepackage{booktabs}
\usepackage{makecell}
\usepackage{warpcol}
\usepackage{enumitem}
\usepackage{algorithm}
\usepackage{algpseudocode}
\usepackage{multirow}
\usepackage{multicol}
\usepackage{mathtools}
\usepackage{xcolor}

\usepackage{color}
\definecolor{citeCol}{RGB}{0,72,187}
\definecolor{linkCol}{RGB}{227,66,77}
\usepackage[pagebackref,breaklinks,colorlinks,urlcolor=magenta,citecolor=citeCol,linkcolor=linkCol,bookmarks=false]{hyperref}

\usepackage{setspace}

\definecolor{amber}{rgb}{1.0, 0.49, 0.0}

\def\eg{\textit{e.g.}}
\def\ie{\textit{i.e.}}
\def\etal{\textit{et al.}}

\def\vs{\textit{vs.}}
\def\wrt{\textit{w.r.t.} }

\usepackage[capitalize]{cleveref}
\crefname{section}{Sec.}{Secs.}
\Crefname{section}{Section}{Sections}
\Crefname{table}{Table}{Tables}
\crefname{table}{Tab.}{Tabs.}

\def\cvprPaperID{1105} 
\def\confName{CVPR}
\def\confYear{2023}

\title{
Neural Preset for Color Style Transfer
}

\author{Zhanghan Ke$^{1}$ \qquad Yuhao Liu$^{1}$ \qquad Lei Zhu$^{1}$ \qquad Nanxuan Zhao$^{2}$ \qquad Rynson W.H. Lau$^{1}$ \\
$^{1}$City University of Hong Kong \qquad $^{2}$Adobe Research \\
{\small Project Page: {\color{magenta}https://zhkkke.github.io/NeuralPreset}}
}

\begin{document}

\twocolumn[{%
\renewcommand\twocolumn[1][]{#1}%
\maketitle

\vspace{-0.9cm}
\begin{center}
    \centering
    \includegraphics[width=1\linewidth]{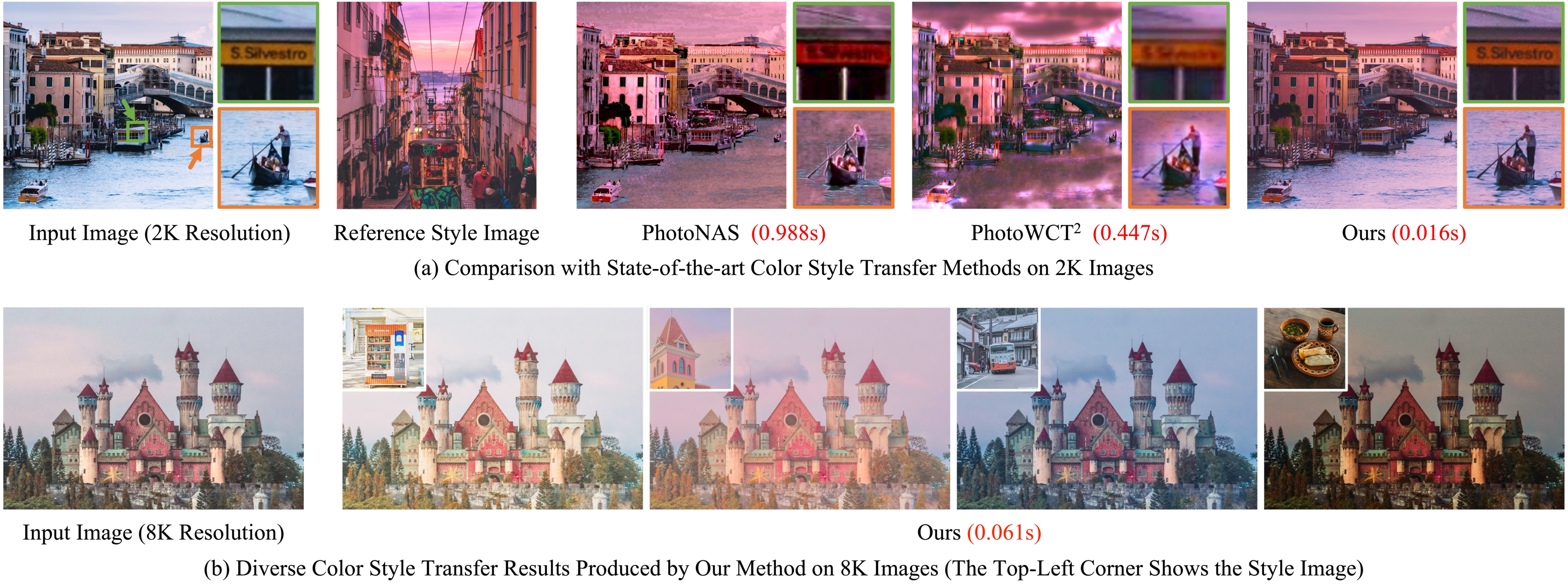}
    \vspace{-0.5cm}
    \captionof{figure}{
    \textbf{Our Color Style Transfer Results.} (a) State-of-the-art methods PhotoNAS~\cite{an2019ultrafast} and PhotoWCT$^{2}$~\cite{PhotoWCT2} produce distort textures (\eg, text in green box) and dissonant colors (\eg, content in orange box). Besides, they have long inference time even on the latest Nvidia RTX3090 GPU (red numbers in brackets). In contrast, our method avoids artifacts and is $\sim$$28\times$ faster. (b) Our method can produce faithful results on 8K images, but both PhotoNAS and PhotoWCT$^{2}$ run into the out-of-memory problem.
    Zoom in for better visualization.
    }
    \vspace{0.3cm}
    \label{fig:quickcomp_new}
\end{center}%
}]

\begin{abstract}
\vspace{-0.3cm}
In this paper, we present a Neural Preset technique to address the limitations of existing color style transfer methods, including visual artifacts, vast memory requirement, and slow style switching speed. Our method is based on two core designs. 
First, we propose Deterministic Neural Color Mapping (DNCM) to consistently operate on each pixel via an image-adaptive color mapping matrix, avoiding artifacts and supporting high-resolution inputs with a small memory footprint.
Second, we develop a two-stage pipeline by dividing the task into color normalization and stylization, which allows efficient style switching by extracting color styles as presets and reusing them on normalized input images.
Due to the unavailability of pairwise datasets, we describe how to train Neural Preset via a self-supervised strategy.
Various advantages of Neural Preset over existing methods are demonstrated through comprehensive evaluations.
Besides, we show that our trained model can naturally support multiple applications without fine-tuning, including low-light image enhancement, underwater image correction, image dehazing, and image harmonization. 
\end{abstract}

\vspace{-0.5cm}
\section{Introduction}
\label{sec:intro}

With the popularity of social media (\eg, Instagram and Facebook), people are increasingly willing to share photos in public. 
Before sharing, color retouching becomes an indispensable operation to help express the story captured in images more vividly and leave a good first impression. 
Photo editing tools usually provide color style presets, such as image filters or Look-Up Tables (LUTs), to help users explore efficiently.
However, these filters/LUTs are handcrafted with pre-defined parameters, and are not able to
generate consistent color styles for images with diverse appearances.
Therefore, careful adjustments by the users is still necessary. 
To address this problem, color style transfer techniques have been introduced
to automatically map the color style from a well-retouched image (\ie, the style image) to another (\ie, the input image).

Earlier color style transfer methods~\cite{ColorTransfer,TransColorToGrey,Piti2005NdimensionalPD,Piti2007AutomatedCG} focus on retouching the input image according to low-level feature statistics of the  style image. 
They disregard high-level information, resulting in unexpected changes in image inherent colors.
Although recent deep learning based models~\cite{DPST,PhotoWCT,WCT2,an2019ultrafast,DeepPreset,PhotoWCT2} give promising results, they typically suffer from three obvious limitations in practice 
(Fig.\,\ref{fig:quickcomp_new}\,(a)).
First, they produce unrealistic artifacts (\eg, distorted textures or inharmonious colors) in the stylized image since they perform color mapping based on convolutional models, which operate on image patches and may have inconsistent outputs for pixels with the same value. 
Although some auxiliary constraints~\cite{DPST} or post-processing strategies~\cite{PhotoWCT} have been proposed, they still fail to prevent artifacts robustly.
Second, they cannot handle high-resolution (\eg, 8K) images due to their huge runtime memory footprint.
Even using a GPU with 24GB of memory, most recent models suffer from the out-of-memory problem when processing 4K images.
Third, they are inefficient in switching styles because they carry out color style transfer as a single-stage process, requiring to run the whole model every time.

In this work, we present a Neural Preset technique with two core designs to overcome the above limitations: 

\textbf{(1)} Neural Preset leverages Deterministic Neural Color Mapping (DNCM) as an alternative to the color mapping process based on convolutional models. By multiplying an image-adaptive color mapping matrix, DNCM converts pixels of the same color to a specific color, avoiding unrealistic artifacts effectively. Besides, 
DNCM operates on each pixel independently with a small memory footprint, supporting very high-resolution inputs. Unlike adaptive 3D LUTs~\cite{zeng2020lut,cvpr2022Cong} that need to regress tens of thousands of parameters or automatic filters~\cite{Harmonizer,Exposure} that perform particular color mappings, DNCM can model arbitrary color mappings with only a few hundred learnable parameters. 

\textbf{(2)} 
Neural Preset carries out color style transfer in two stages to enable fast style switching.
Specifically, the first stage builds a \textit{nDNCM} from the input image for color normalization, which maps the input image to a normalized color style space representing the ``image content''; the second stage builds a \textit{sDNCM} from the style image for color stylization, which transfers the normalized image to the target color style.
Such a design has two advantages in terms of efficiency: 
the parameters of \textit{sDNCM} can be stored as color style presets and reused by different input images, while the input image can be stylized by diverse color style presets after normalized once with \textit{nDNCM}.

In addition, since there are no pairwise datasets available, we propose a new self-supervised strategy for Neural Preset to be trainable. Our comprehensive evaluations demonstrate that Neural Preset outperforms state-of-the-art methods significantly in various aspects.
Notably, Neural Preset can produce faithful results for 8K images (Fig.\,\ref{fig:quickcomp_new}\,(b)) 
and can provide consistent color style transfer results across video frames without post-processing. 
Compared to recent deep learning based models, Neural Preset achieves $\sim$$28\times$ speedup on a Nvidia RTX3090 GPU, supporting real-time performances at 4K resolution without engineering tricks like model quantization. 
Finally, we show that our trained model can be applied to other color mapping tasks without fine-tuning, including low-light image enhancement~\cite{low-light-survey}, underwater image correction~\cite{underwater-survey}, image dehazing~\cite{dehazing-survey}, and image harmonization~\cite{harmonization-survey}.

\section{Related Works}

\noindent\textbf{Color Style Transfer.}\quad
Unlike artistic style transfer~\cite{Li2016CombiningMR,Johnson2016Perceptual,d135f1128a,Chen2017StyleBankAE,huang2017adain,Dumoulin2017ALR,deng2021stytr2,DEAIST,Hong_2021_ICCV,Cheng_2021_CVPR,Gatys2016ImageST} that alters both textures and colors of images, color style transfer ({\it aka} photorealistic style transfer) aims to shift only the colors from one image to another. 
Traditional methods~\cite{ColorTransfer,TransColorToGrey,Piti2005NdimensionalPD,Piti2007AutomatedCG} mostly match the statistics of low-level features, such as the mean and variance of images~\cite{ColorTransfer} or the histograms of filter responses~\cite{Piti2005NdimensionalPD}. However, these methods often transfer unwanted colors if the style and input images have large appearance differences.
Recently, many methods exploiting convolutional neural networks (CNNs)~\cite{DPST,PhotoWCT,WCT2,an2019ultrafast,DeepPreset,PhotoWCT2} are proposed for color style transfer.
For example, Yoo \etal~\cite{WCT2} introduce a model with wavelet pooling/unpooling to reduce distortions. 
An \etal~\cite{an2019ultrafast} use network architecture search to explore a more effective asymmetric model.
Chiu \etal~\cite{PhotoWCT2} propose to obtain a more compact model by block-wise training with a coarse-to-fine transformation.
To address the limitations (\ie, visual artifacts, huge memory consumption, and inefficient in switching styles) of the aforementioned methods as stated in Sec.\,\ref{sec:intro}, we present Neural Preset that supports artifact-free color style transfer with only a small memory footprint, via DNCM, and enables fast style switching, via a two-stage pipeline.

\medskip
\noindent\textbf{Deterministic Color Mapping with CNNs.}\quad
Filters and LUTs avoid artifacts as they perform \textit{deterministic color mapping} to produce consistent outputs for the same input pixel values.
Some recent image enhancement and harmonization methods~\cite{zeng2020lut,cvpr2022Cong,Harmonizer,Exposure} have attempted to implement color mapping using filters/LUTs with image-adaptive parameters predicted by CNNs. 
However, combining filters/LUTs with CNNs for color mapping has clear drawbacks.
Filter-based methods~\cite{Harmonizer,Exposure} integrate a finite number of image filters, and can only handle basic color adjustments, \eg, brightness and contrast.
LUT-based methods~\cite{zeng2020lut,cvpr2022Cong} need to predict coefficients to linearly merge several template LUTs, 
because LUTs have a large number of learnable parameters that are difficult to optimize via CNNs.
Although the affine bilateral grid~\cite{hdrnet} may model complex color mapping with fewer learnable parameters, it cannot provide deterministic color mapping. 
Applying it to color style transfer~\cite{JBL} may lead to inharmonious colors in different regions of the image.
Instead of adopting the aforementioned schemes, we propose DNCM that has only a few hundred learnable parameters but can model arbitrary deterministic color mappings.

\medskip
\noindent\textbf{Self-Supervised Learning (SSL).}\quad
SSL has been widely explored for pre-training~\cite{7780647,Chen2020GenerativePF,MAASVL,URLPRL,ULVR,LFWOM,MOCO,simclr,BYOL}. 
Some works also solve specific vision tasks via SSL~\cite{Hoyer2021ThreeWT,Laine2019HighQualitySD,SSH}.
Since it is expensive to annotate ground truths for color style transfer,
most methods either minimize perceptual losses~\cite{Gatys2016ImageST,Chen2017StyleBankAE,huang2017adain,DPST} or match the statistics of image features~\cite{WCT,PhotoWCT,WCT2,PhotoWCT2}.
However, such weak constraints usually result in severe visual artifacts. Yim \etal~\cite{FST} and Ho \etal~\cite{DeepPreset} suggest imposing stronger constraints by reconstructing perturbed images, but their trained models can only transfer color styles to images with natural appearances, \eg, images taken by a camera without post-processing. 
To this end, we present a new SSL strategy for Neural Preset, which not only learns from reconstructing perturbed images but also enables the trained models to transfer color styles between arbitrary images.

\section{Method}
Our Neural Preset performs color style transfer through a two-stage pipeline, where both stages employ DNCM for color mapping. 
In this section, we first introduce DNCM in details (Sec.\,\ref{sec:ncm}).
We then present the two-stage DNCM-based color style transfer pipeline (Sec.\,\ref{sec:pipeline}). Finally, we describe our self-supervised learning strategy for training Neural Preset (Sec.\,\ref{sec:ssl}).

\subsection{Deterministic Neural Color Mapping (DNCM)}\label{sec:ncm}

A straightforward idea to model deterministic color mapping that can adapt to different images is to combine filters/LUTs with the image-adaptive parameters predicted by CNNs.
However, each image filter can only provide a single color mapping. Integrating a finite number of filters can only cover a limited range of color mappings.
Besides, as a common 32-level 3D LUT has $\sim$10K parameters, it is infeasible to regress image-specific 3D LUTs. The approach based on predicting coefficients to merge template LUTs still needs to optimize tens of thousands of parameters to build template LUTs.

 \begin{figure}[t]
\centering
\includegraphics[width=0.99\linewidth]{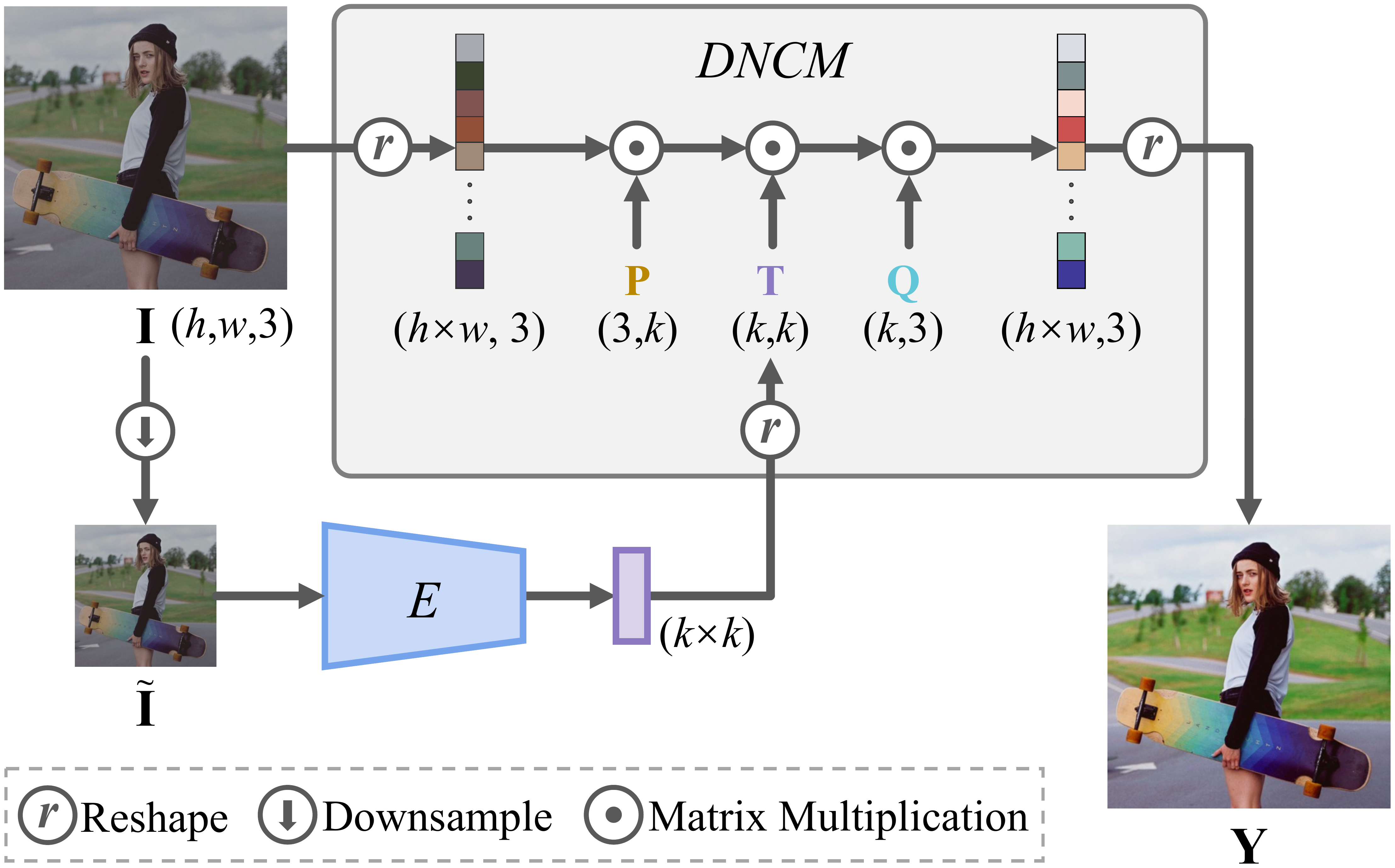}
{\begin{center}
\vspace{-0.3cm}
\caption{\textbf{Illustration of DNCM.}
The DNCM parameters consist of two color projection matrices ($\mathbf{P}$, $\mathbf{Q}$) and image-adaptive parameters $\mathbf{T}$ predicted by an encoder $E$. Here, we obtain $\mathbf{\tilde{I}}$ by downsampling $\mathbf{I}$.
DNCM maps the input image $\mathbf{I}$ to output image $\mathbf{Y}$ by multiplying $\mathbf{I}$ with $\mathbf{P}$, $\mathbf{T}$, and $\mathbf{Q}$ sequentially. 
}
\label{fig:ncm}
\end{center}
}
\vspace{-0.7cm}
\end{figure}

 \begin{figure*}[t]
\centering
\includegraphics[width=0.99\linewidth]{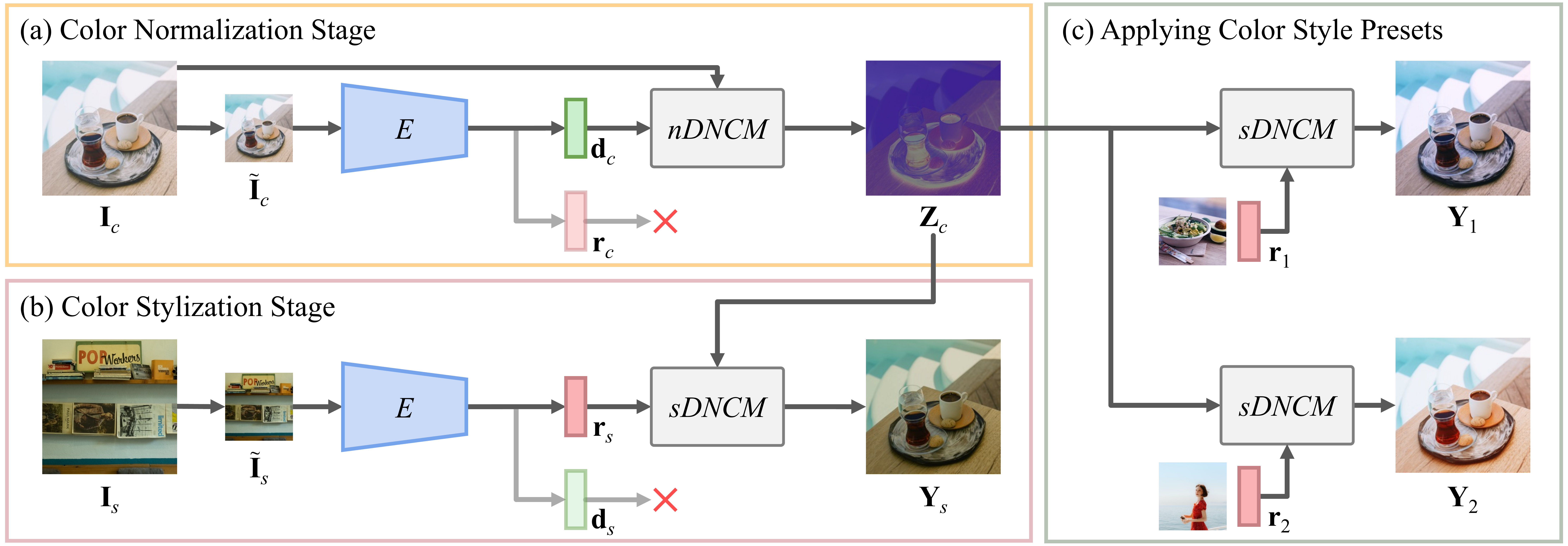}
{\begin{center}
\vspace{-0.3cm}
\caption{\textbf{Overview of Our Pipeline.} Our pipeline consists of two stages: (a) in the first stage, the input image $\textbf{I}_{c}$ is converted to an image $\textbf{Z}_{c}$ in the normalized color style space via \textit{nDNCM} with parameters $\textbf{d}_{c}$; (b) in the second stage, the color style parameters $\textbf{r}_{s}$ are extracted from the style image $\textbf{I}_{s}$ for \textit{sDNCM} to map $\textbf{Z}_{c}$ to $\textbf{Y}_{s}$, which will then have the same color style as $\textbf{I}_{s}$. Besides, the design of our pipeline supports fast style switching: in 
(c), the preset color style parameters $\textbf{r}_{1}$/$\textbf{r}_{2}$ can be reused by \textit{sDNCM} to stylize $\textbf{Z}_{c}$ to obtain $\textbf{Y}_{1}$/$\textbf{Y}_{2}$.}
\label{fig:pipeline}
\end{center}
}
\vspace{-0.7cm}
\end{figure*}

Here, we propose DNCM to model arbitrary deterministic color mapping with much fewer learnable parameters.
As shown in Fig.\,\ref{fig:ncm}, given an input image $\mathbf{I}$ of size $(h, w, 3)$, we downsample it 
to obtain a thumbnail $\mathbf{\tilde{I}}$, which provides image-adaptive color mapping matrix $\mathbf{T}$ for DNCM.
Specifically, we feed $\mathbf{\tilde{I}}$ into an encoder $E$ to predict $\mathbf{T}$ of size $(k \times k)$, and then reshape it to a size of $(k,k)$, as:
\begin{equation}\label{eq:ncm_T}
     \mathbf{T}^{(k \times k)} = E(\,\mathbf{\tilde{I}}\,), \;\; \mathbf{T}^{(k \times k)}
      \rightarrow 
      \mathbf{T}^{(k,k)},
\end{equation}
where $\rightarrow$ denotes the reshape operation, and $k$ is  empirically set to a small value (\eg, 16).
With matrix $\mathbf{T}^{(k, k)}$, we form DNCM to alter the colors of $\mathbf{I}$. In DNCM, we first unfold $\mathbf{I}$ as a 2D matrix of size $(h \times w, 3)$. We then embed each pixel in $\mathbf{I}$ into a \textit{k}-dimensional vector by a projection matrix $\mathbf{P}^{(3, k)}$.
After that, we multiply the embedded vectors by $\mathbf{T}^{(k, k)}$.
Finally, we apply another projection matrix $\mathbf{Q}^{(k, 3)}$ to convert the embedded vectors back to the RGB color space and reshape pixels to output $\mathbf{Y}$ with new colors. 
Note that both $\mathbf{P}$ and $\mathbf{Q}$ are learnable matrices shared by all images.
Formally, DNCM can be defined as:
\begin{equation}\label{eq:ncm_Y}
    \mathbf{Y} 
    = \textit{DNCM}(\mathbf{I} , \, \mathbf{T}) 
    = \mathbf{I}^{(h \times w, 3)} \cdot \mathbf{P}^{(3,k)} \cdot \mathbf{T}^{(k,k)} \cdot \mathbf{Q}^{(k,3)}, 
\end{equation}
where $\cdot$ denotes matrix multiplication.
We omit the reshape operations in Eq.\,\ref{eq:ncm_Y} for simplicity.
Note that $\mathbf{Q}$ is not the inverse of $\mathbf{P}$, since $\mathbf{P}$ is not a square matrix when $k \neq 3$.

Implementing color mapping as
Eq.\,\ref{eq:ncm_Y} brings three main benefits.
First, it effectively avoids visual artifacts as pixels of the same color in $\mathbf{I}$ will still have the same color after being mapped to $\mathbf{Y}$.
Second, it requires only a small memory footprint since each pixel is processed independently with efficient matrix multiplications.
Third, it makes $E()$ easy to optimize as only $k \times k$ image-adaptive parameters (\ie, $\mathbf{T}$) should be regressed.

\subsection{Two-Stage Color Style Transfer Pipeline}\label{sec:pipeline}

Recent methods implicitly embed the color style transfer process into CNNs to form single-stage pipelines. Therefore, they must run the whole pipeline every time to compute the color mapping between two images, which is inefficient when applying diverse color styles to an image or transferring a color style to multiple images.

In contrast, we design an explicit two-stage pipeline based on DNCM.
The key insight behind our pipeline is that if we can separate the color style of an image from its ``image content'', we can effectively transfer different color styles
to the ``image content''. 
However, to achieve this, we need to answer two questions.
First, how to remove or add color styles? Since we need to alter the image color style but preserve the ``image content'', we propose to utilize a pair of \textit{nDNCM} and \textit{sDNCM}.
While \textit{nDNCM} converts the input image to 
a space
that contains only the ``image content'', \textit{sDNCM} 
transfers the 
“image content”
to the target color style, 
using parameters extracted from the style image.
Second, how to represent ``image content''? Since this concept is difficult to define through hand-crafted features, we propose to learn a normalized color style space representing the ``image content'' by back-propagation. In such a normalized color style space,
images of the same content but with different color styles should have a consistent appearance, \ie, the same normalized color style.

As shown in Fig.\,\ref{fig:pipeline}\,(a)(b), we modify the encoder $E$ to output $\mathbf{d}$ and $\mathbf{r}$, which are applied as the parameters of \textit{nDNCM} and \textit{sDNCM}, respectively. 
Suppose that we want to transfer the color style of a style image $\mathbf{I}_{s}$ to an input image $\mathbf{I}_{c}$. 
In the first stage, we convert $\mathbf{I}_{c}$ to $\mathbf{Z}_{c}$ in the normalized color style space
via \textit{nDNCM} with $\mathbf{d}_{c}$ predicted from $\mathbf{\tilde{I}}_{c}$, as: 
\begin{equation}\label{eq:pipeline_d}
    \mathbf{Z}_{c} = \textit{nDNCM}(\mathbf{I}_{c}, \, \mathbf{d}_{c}),
    \;\; \text{where} \;\;
    \{\mathbf{d}_{c}, \mathbf{r}_{c}\} = E(\mathbf{\tilde{I}}_{c}).
\end{equation}
In the second stage, we extract $\mathbf{r}_{s}$, \ie, the parameters containing the color style of $\mathbf{I}_{s}$, to transfer $\mathbf{Z}_{c}$ to the stylized image $\mathbf{Y}_{s}$ via \textit{sDNCM}, as:
\begin{equation}\label{eq:pipeline_r}
    \mathbf{Y}_{s} = \textit{sDNCM}(\mathbf{Z}_c, \, \mathbf{r}_{s}), 
    \;\; \text{where} \;\;
    \{\mathbf{d}_{s}, \mathbf{r}_{s}\}  = E(\mathbf{\tilde{I}}_{s}).
\end{equation}
$E()$ used in Eq.\,\ref{eq:pipeline_d} and \ref{eq:pipeline_r} have shared weights. \textit{nDNCM} and \textit{sDNCM} have different projection matrices $\mathbf{P}$ and $\mathbf{Q}$.

By storing color style parameters (\eg, $\mathbf{r}_{s}$) as presets and reusing them to construct \textit{sDNCM}, our pipeline can support fast style switching using color style presets --
we only need to normalize the color style of the input image once, and can then quickly retouch it to diverse color styles by \textit{sDNCM} with the stored presets (\eg, $\mathbf{r}_{s}$). 
For example, in Fig.\,\ref{fig:pipeline}\,(c), 
we apply presets $\mathbf{r}_{1}$ and $\mathbf{r}_{2}$ on $\mathbf{Z}_{c}$ to obtain $\mathbf{Y}_{1}$ and $\mathbf{Y}_{2}$.

\subsection{Self-Supervised Training Strategy}\label{sec:ssl}

 \begin{figure}[t]
 \vspace{0.1cm}
\centering
\includegraphics[width=0.99\linewidth]{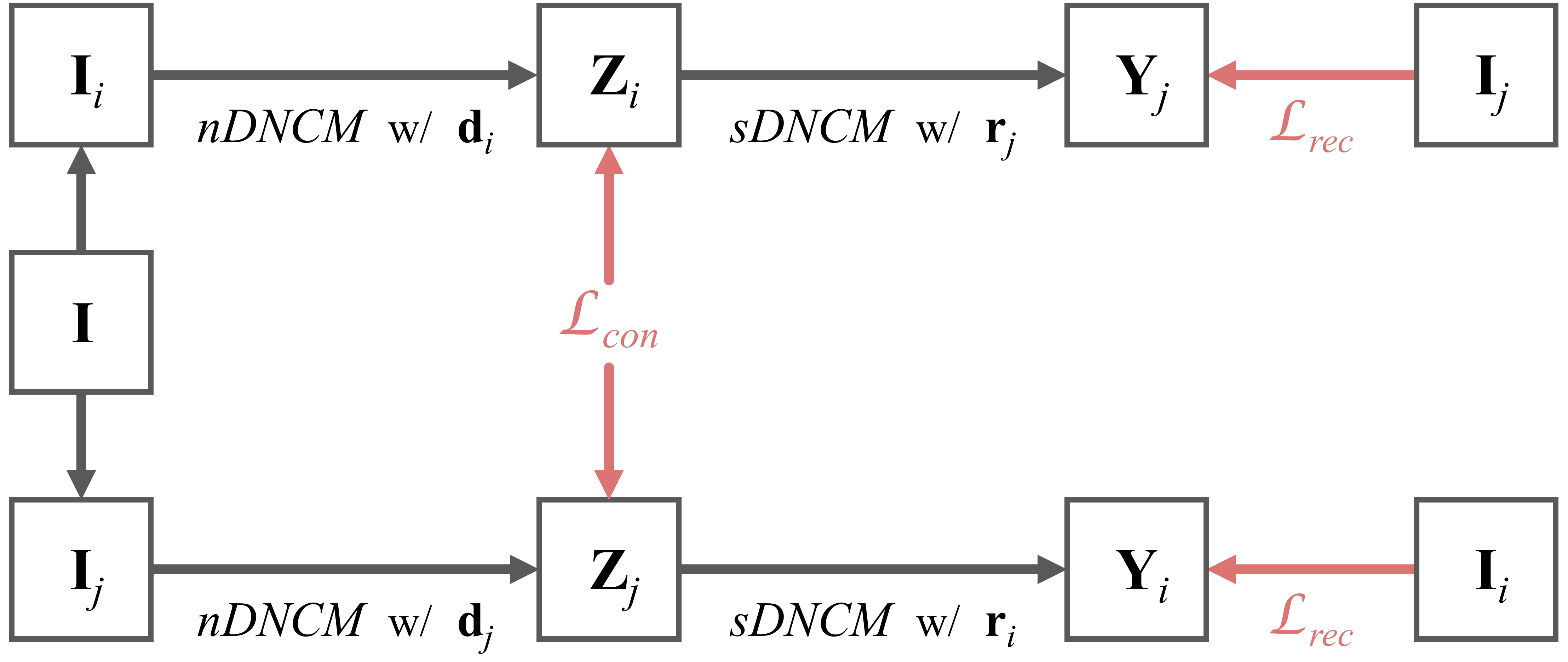}
{\begin{center}
\vspace{-0.3cm}
\caption{\textbf{Our Self-Supervised Training Strategy.} Both $\mathbf{I}_{i}$/$\mathbf{I}_{j}$ are generated from $\mathbf{I}$ via random color perturbations. We constrain $\mathbf{Z}_{i}$/$\mathbf{Z}_{j}$ to be the same via a consistency loss $\mathcal{L}_{con}$ and learn style transfer results $\mathbf{Y}_{i}$/$\mathbf{Y}_{j}$ via a reconstruction loss $\mathcal{L}_{rec}$.}
\label{fig:ssl}
\end{center}
}
\vspace{-0.7cm}
\end{figure}

We develop a self-supervised strategy to train Neural Preset, as shown in Fig.\,\ref{fig:ssl}.
Since no ground truth stylized images are available, we create pseudo stylized images from the input image $\mathbf{I}$. Specifically, we add perturbations on $\mathbf{I}$ to obtain two augmented samples with different color styles, which are denoted as $\mathbf{I}_{i}$ and $\mathbf{I}_{j}$. The perturbations we use involve operations that only change image colors, \eg, random image filters or LUTs. 

The first stage of our pipeline aims to normalize the color style of input images, which means that input images with the same content but different color styles should be consistent in the normalized color style space. Hence, we apply a L2 consistency loss between the outputs of this stage. 
Formally, we predict the \textit{nDNCM} parameters $\mathbf{d}_{i}$/$\mathbf{d}_{j}$ to transfer $\mathbf{I}_{i}$/$\mathbf{I}_{j}$ to $\mathbf{Z}_{i}$/$\mathbf{Z}_{j}$, 
and we constrain $\mathbf{Z}_{i}$/$\mathbf{Z}_{j}$ by: 
\begin{equation}\label{eq:Lcon}
\begin{split}
        \mathcal{L}_{con} & = || \mathbf{Z}_{i} - \mathbf{Z}_{j} ||_{2} \\
        & = ||\textit{nDNCM}(\mathbf{I}_{i}, \, \mathbf{d}_{i}) - \textit{nDNCM}(\mathbf{I}_{j}, \, \mathbf{d}_{j})||_{2}.
\end{split}
\end{equation}
The second stage of our pipeline aims to stylize the normalized images. 
To convert the input images to new styles, we swap the predicted \textit{sDNCM} parameters $\mathbf{r}$ to stylize the two samples, \ie, $\mathbf{Z}_{i}$ will be stylized by $\mathbf{r}_{j}$ while $\mathbf{Z}_{j}$ by
$\mathbf{r}_{i}$, as:
\begin{equation}
     \mathbf{Y}_{i} = \textit{sDNCM}(\mathbf{Z}_{j}, \, \mathbf{r}_{i}), \quad \mathbf{Y}_{j} = \textit{sDNCM}(\mathbf{Z}_{i}, \, \mathbf{r}_{j}).
\end{equation}
We apply a L1 reconstruction loss between $\mathbf{I}$ and $\mathbf{Y}$ to learn color style transfer, as:
\begin{equation}\label{eq:Lrec}
        \mathcal{L}_{rec} = || \mathbf{Y}_{i} - \mathbf{I}_{i} ||_{1} + || \mathbf{Y}_{j} - \mathbf{I}_{j} ||_{1}.
\end{equation}
The final loss is a combination of Eq.\,\ref{eq:Lcon} and \ref{eq:Lrec}, as:
\begin{equation}\label{eq:L}
    \mathcal{L} = \mathcal{L}_{rec} + \lambda \, \mathcal{L}_{con},
\end{equation}
where $\lambda$ is a controllable weight.
Refer to the analysis in 
Appendix\,\ref{appendix:optimization} 
on how to derive our training constraints from the fundamental color style transfer objective.

\section{Experiments}
In this section, we first introduce our experimental setting, including datasets, implementation, and quantitative metrics. We then extensively compare Neural Preset with existing color style transfer methods (Sec.\,\ref{sec:comparision}). We further analyze the components and hyper-parameters of Neural Preset through ablation experiments (Sec.\,\ref{sec:ablation}). 

\medskip
\noindent\textbf{Datasets.}\quad 
Following recent color style transfer methods~\cite{WCT2,an2019ultrafast,PhotoWCT}, we train our model on the images from the MS COCO~\cite{MSCOCO} dataset.
We use about 5,000 LUT files, along with the random image filter adjustment strategy~\cite{Harmonizer}, as input color perturbations during training. 
We collect 50 images with diverse color styles and pair each two of them to build a validation set consisting of 2,500 samples.

\medskip
\noindent\textbf{Implementation.}\quad 
We adopt EfficientNet-B0~\cite{EfficientNet} as the encoder $E$ in Neural Preset.
We fix the input size of $E$ to $256 \times 256$.
We set the parameter dimension $k$ in DNCM to $16$, so the number of parameters predicted by $E$ is only $256$ for each image. 
We train Neural Preset by the Adam~\cite{Adam} optimizer for $32$ epochs. With a batch size of $24$, the initial learning rate is $3e^{-4}$ and is multiplied by $0.1$ after $24$ epochs. We set the loss weight $\lambda$ to $10$. 

 \begin{figure*}[t]
\centering
\includegraphics[width=0.99\linewidth]{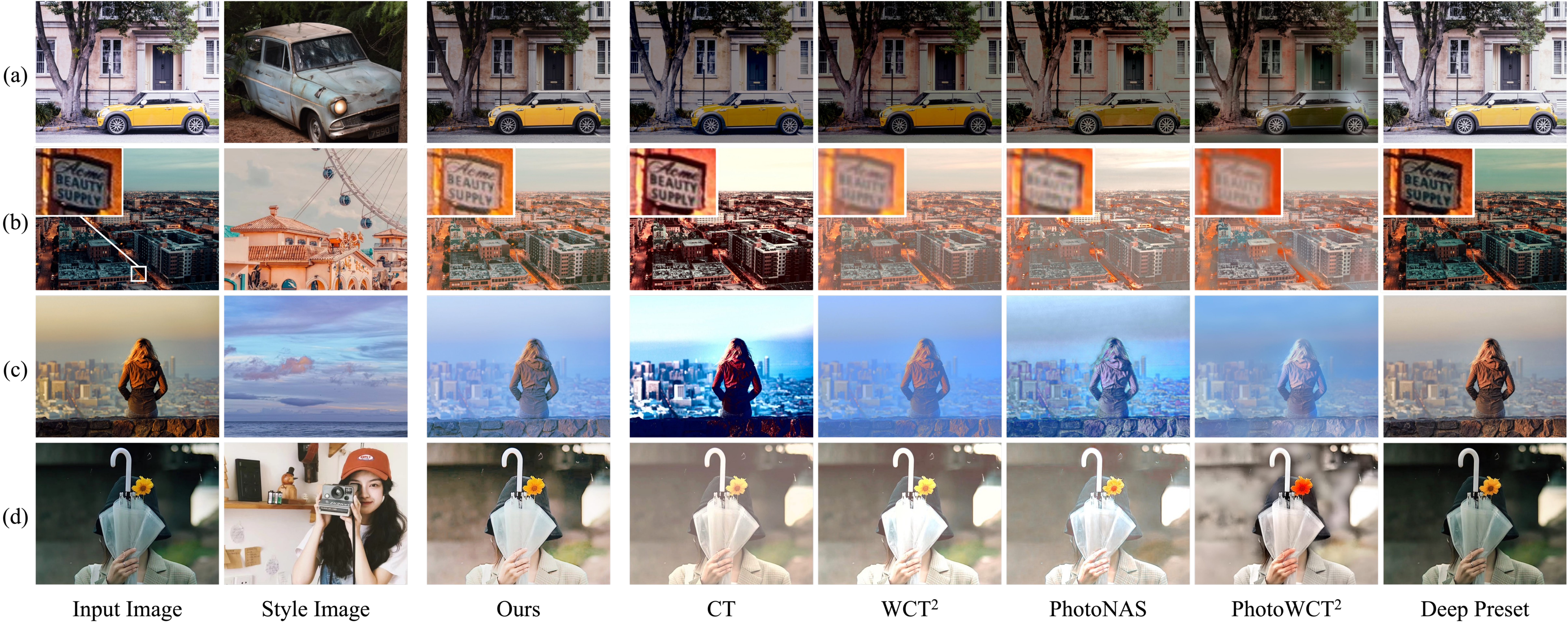}
{\begin{center}
\vspace{-0.3cm}
\caption{\textbf{Qualitative Comparison.} Our method has advantages in (a) producing natural stylized images, (b) preserving image textures, 
(c) maintaining object inherent colors, and (d) providing color properties more consistent with the style images. 
}
\label{fig:visualcomp}
\end{center}
}
\vspace{-0.5cm}
\end{figure*}

\begin{table*}[th]
  \begin{center}
\setlength{\tabcolsep}{6pt}
\small
\begin{tabular}{r|ccccccc}
\toprule
Method & CT~\cite{ColorTransfer} & PhotoWCT~\cite{PhotoWCT} & WCT$^{2}$~\cite{WCT2} & PhotoNAS~\cite{an2019ultrafast} & PhotoWCT$^{2}$~\cite{PhotoWCT2} & Deep Preset~\cite{DeepPreset} & Ours  \\
\midrule
Average Ranking $\downarrow$ & 4.97 & 5.75 &  2.67 & 3.39 & 4.11 & 5.30 & \textbf{1.81} \\

\bottomrule
\end{tabular}
    \caption{\textbf{Average Ranking of Different Methods in the User Study.} The lower the number, the better the human subjective evaluation.
    }\label{tab:userstudy}
\vspace{-0.4cm}
\end{center}
\end{table*}

\medskip
\noindent\textbf{Quantitative Metrics.}\quad
We follow prior works~\cite{WCT2,an2019ultrafast,JBL} to quantitatively evaluate color style transfer quality in terms of style similarity (between the output and the reference style image) and content similarity (between the output and the input image).
However, we observe from our experiments that the metrics used by prior works cannot reflect color style transfer quality precisely.
Therefore, we propose improved style/content similarity measures as our quantitative metrics.
For style similarity measure, prior works use the VGG~\cite{VGG} features learned from ImageNet~\cite{ImageNet} to compute the Gram metric. Since the VGG features are not designed for comparing color styles and contain semantic information, the style similarity that they produce may not be accurate due to semantic bias.
Instead, we train a discriminator~\cite{GAN} model on an annotated dataset (containing 700\,+ color style categories, each with 6-10 images of the same color style retouched by human experts) to accurately predict the color style similarity score (in $[0,1]$) of two images.
For content similarity measure, prior works compute the SSIM metric based on image edges extracted by an edge detection model HED~\cite{HN_Edge}. However, HED  only predicts rough edges and is often incorrect. Hence, we replace HED with a recent model LDC~\cite{xsoria2022ldc}, which can output fine and correct edges for SSIM calculation, providing a  reliable content similarity score (in $[0,1]$).
Refer to 
Appendix\,\ref{appendix:quantitive_eval} 
for more details on our improved metrics.

\subsection{Comparisons}\label{sec:comparision}

We compare Neural Preset with deep learning  based methods (PhotoWCT\,\cite{PhotoWCT}, WCT$^{2}$\,\cite{WCT2}, PhotoNAS\,\cite{an2019ultrafast}, PhotoWCT$^{2}$\,\cite{PhotoWCT2}, and Deep Preset\,\cite{DeepPreset}) as well as a traditional method (CT~\cite{ColorTransfer}).
We evaluate the pre-trained models released by their authors. 
We do not compare with methods whose codes, models, and demos are all unavailable.

\medskip
\noindent\textbf{Qualitative Results.}\quad
Fig.\,\ref{fig:visualcomp} shows the superiority of our qualitative results.
First, Neural Preset produces more natural stylized images (\eg, the colors of the car and wall in Fig.\,\ref{fig:visualcomp}\,(a)). 
Second, Neural Preset can preserve fine textures with target color styles (\eg, the enlarged text in Fig.\,\ref{fig:visualcomp}\,(b)).
Third, Neural Preset is better at maintaining the inherent object colors (\eg, the human hair and clothing regions in Fig.\,\ref{fig:visualcomp}\,(c)). 
Fourth, the outputs of Neural Preset have more consistent color properties with the style images (\eg, the brightness and contrast in Fig.\,\ref{fig:visualcomp}\,(d)). 
The results of PhotoWCT are omitted in Fig.\,\ref{fig:visualcomp} as its visual results are similar to PhotoWCT$^2$.
Refer to 
Appendix\,\ref{appendix:image_results} 
for more visual results of Neural Preset.
Refer to 
Appendix\,\ref{appendix:video_results} 
for video stylization results of Neural Preset.

\begin{figure}[t]
 \vspace{0.1cm}
\centering
\includegraphics[width=0.99\linewidth]{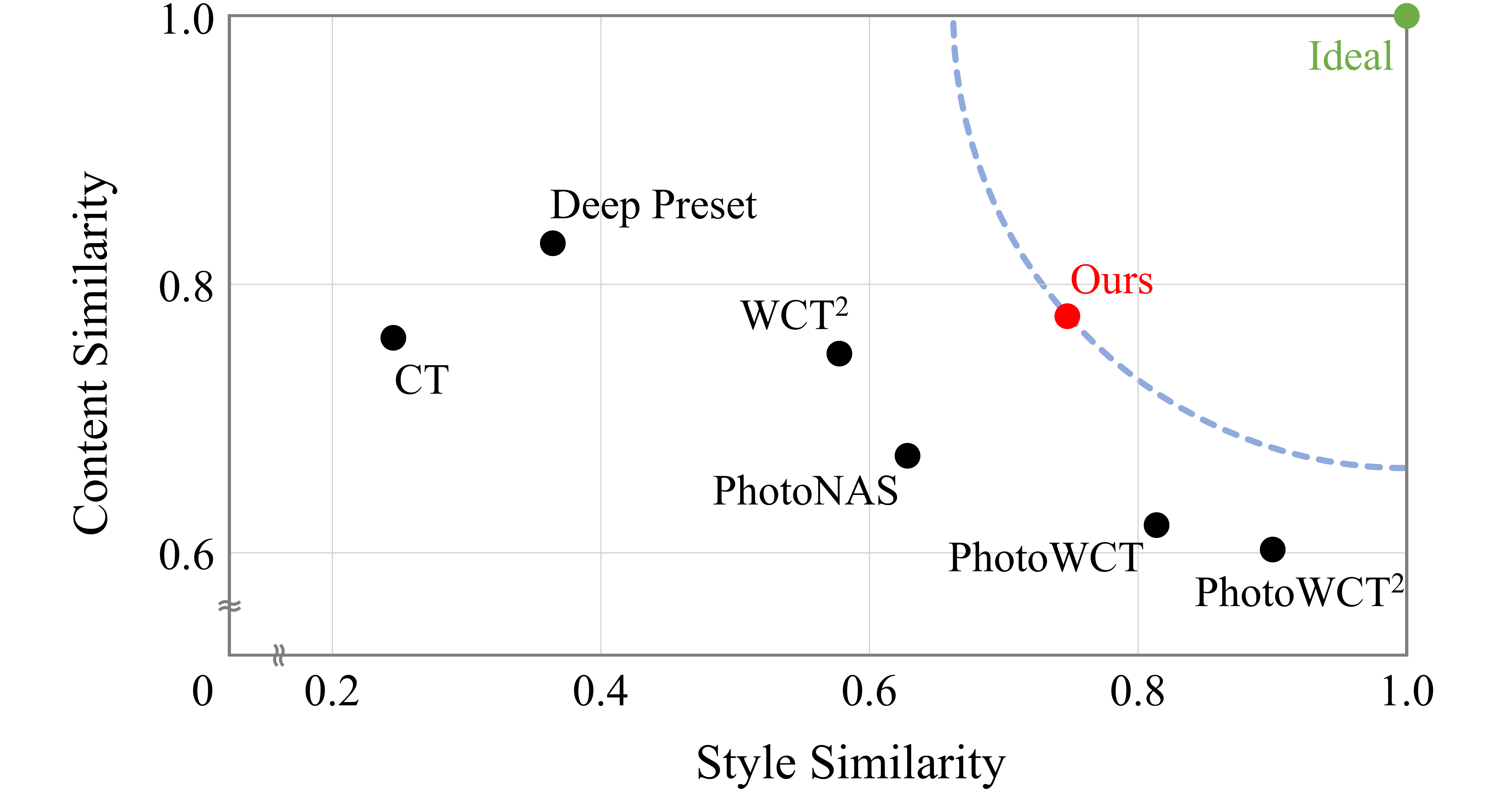}
{\begin{center}
\vspace{-0.3cm}
\caption{\textbf{Quantitative Comparison.} The higher the content similarity and the style similarity, the better the stylization quality. ``Ideal'' refers to the best possible quality. The points on the blue dashed curve have equal distance from ``Ideal'' as our method, \ie, have equivalent color style transfer quality as our method.}
\label{fig:performance}
\end{center}
}
\vspace{-0.7cm}
\end{figure}

\medskip
\noindent\textbf{Quantitative Results.}\quad 
Fig.\,\ref{fig:performance} shows that our Neural Preset comes closest to the ``Ideal'' stylization quality. Although PhotoWCT$^{2}$ has high style similarity scores in Fig.\,\ref{fig:performance}, we can observe from Fig.\,\ref{fig:visualcomp} that it tends to overfit the color styles of the reference images, which however leads to worse visual results. Deep Preset has high content similarity scores in Fig.\,\ref{fig:performance} since it often fails to alter the color style of input images, as shown in Fig.\,\ref{fig:visualcomp}.
Besides, both scores for CT are low as it sometimes produces very erratic  results.

\begin{table*}[t]
  \begin{center}
\setlength{\tabcolsep}{9pt}
\small
\begin{tabular}{l|ccccr}
\toprule
\multirow{2}{*}{Method} & \multicolumn{4}{c}{GPU Inference Time $\downarrow$ / Memory $\downarrow$} & \multicolumn{1}{c}{Model Size $\downarrow$} \\
\cmidrule(lr){2-5} \cmidrule(l){6-6}
& \makecell[c]{FHD \\ (1920 $\times$ 1080)} & \makecell[c]{2K \\ (2560 $\times$ 1440)} & \makecell[c]{4K \\ (3840 $\times$ 2160)} & \makecell[c]{8K \\ (7680 $\times$ 4320)} & \makecell[r]{Number of \\ Parameters}   \\
\midrule
PhotoWCT~\cite{PhotoWCT} & 0.599\,s / 10.00\,GB & 1.002\,s / 16.41\,GB & OOM & OOM & \;\;8.35\,M \\
WCT$^{2}$~\cite{WCT2} & 0.557\,s / 18.75\,GB & OOM & OOM & OOM & 10.12\,M  \\
PhotoNAS~\cite{an2019ultrafast} & 0.580\,s / 15.60\,GB & 0.988\,s / 23.87\,GB & OOM & OOM & 40.24\,M  \\
Deep Preset~\cite{DeepPreset} & 0.344\,s / \;\;8.81\,GB & 0.459\,s / 13.21\,GB & 
1.128\,s / 22.68\,GB & OOM & 267.77\,M  \\
PhotoWCT$^{2}$~\cite{PhotoWCT2} & 0.291\,s / 14.09\,GB & 0.447\,s / 19.75\,GB &  1.036\,s / 23.79\,GB & OOM & 7.05\,M  \\
Ours & \textbf{0.013\,s} / \;\;\textbf{1.96\,GB} & \textbf{0.016\,s} / \;\;\textbf{1.96\,GB} & \textbf{0.019\,s} / \;\;\textbf{1.96\,GB} & \textbf{0.061\,s} / \textbf{1.96\,GB} & \textbf{5.15\,M} \\
\bottomrule
\end{tabular}
\caption{\textbf{Comparison on GPU Inference Time\,/\,Memory, and Model Size.} 
    All evaluations are conducted with Float32 model precision on a Nvidia RTX3090 GPU (24GB memory). 
    The values in parentheses under the resolutions are the exact image width and height. The units ``s'', ``GB'', and ``M'' mean seconds, gigabytes, and millions, respectively.
    ``OOM'' means having the out-of-memory issue.
    }\label{tab:speed_size}
\vspace{-0.4cm}
\end{center}
\end{table*}

\medskip
\noindent\textbf{User Study.}\quad
We further conduct a user study to evaluate the subjective quality of different methods.
We invite 58 users and show them 20 image sets randomly selected from our validation set, with each image set consisting of an input image, a reference style image, and 7 randomly shuffled color style transfer results. 
For each image set, the users are required to rank the overall stylization quality of the 7 results by considering the style/content similarity and the photorealism of the results as well as whether the color style of the results is visually pleasing. 
After collected 1,160 (58$\times$20) results, we compute the average ranking of each method.
Table\,\ref{tab:userstudy} shows that results from our method 
are largely preferred by the users. 
Note that the second-ranked WCT$^2$ can only handle images of FHD resolution (see Table\,\ref{tab:speed_size}). 
We provide Top1-Top3 ratios 
Appendix\,\ref{appendix:more-user-study_results}.

\medskip
\noindent\textbf{Inference Efficiency and Model Size.}\quad
As shown in Table\,\ref{tab:speed_size}, Neural Preset achieves nearly $28\times$ speedup compared to the fastest state-of-the-art method (\ie, PhotoWCT$^{2}$~\cite{PhotoWCT2}) on 2K images.
Neural Preset also enables real-time inference (about $52$\,fps) at 4K resolution, and can handle 8K resolution images at over $16$\,fps.
Refer to 
Appendix\,\ref{appendix:compare_cpu_speed} 
for the inference time on CPU. 
Table\,\ref{tab:speed_size} also shows that existing methods require large amounts of memory for inference. Using even a GPU with 24GB memory, many of them still have the out-of-memory problem at 4K resolution, and all of them fail at 8K resolution.
In contrast, Neural Preset requires only 1.96GB of memory, irrespective of the image resolution.
This is because \textit{nDNCM}/\textit{sDNCM} in Neural Preset operate on each pixel independently, allowing us to save memory via splitting high-resolution images into small patches for processing. 
Besides, Neural Preset also has the lowest number of parameters.


\begin{figure}[h]
\centering
\includegraphics[width=0.99\linewidth]{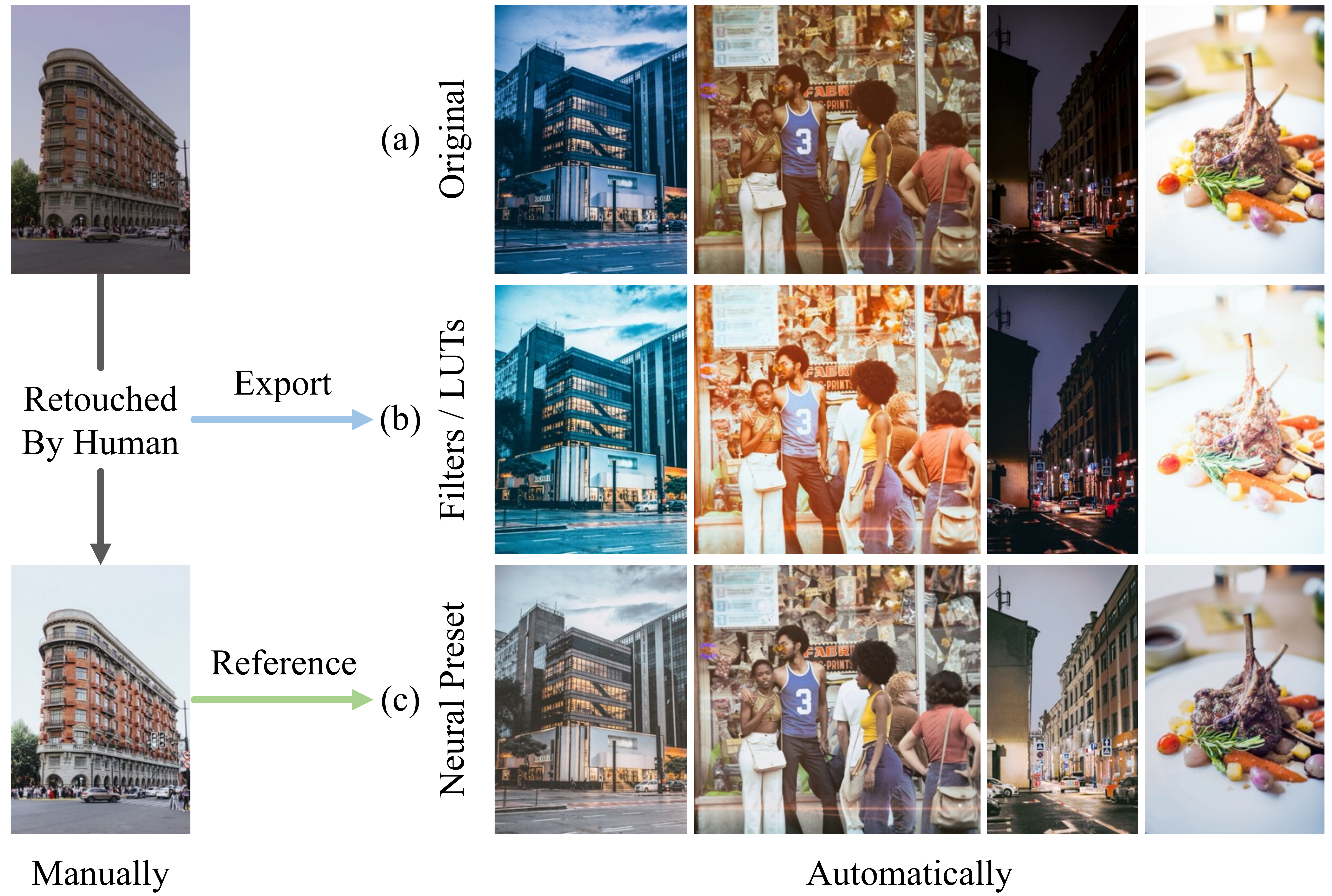}
{\begin{center}
\vspace{-0.3cm}
\caption{\textbf{Comparing Neural Preset with Filters/LUTs.} (c) The results of Neural Preset have more consistent color styles than (b) the results of filters/LUTs.
}
\label{fig:filterlutcomp}
\end{center}
}
\vspace{-0.7cm}
\end{figure}


\medskip
\noindent\textbf{Comparing Neural Preset with Filters and Luts.}\quad
After manually retouching an image by a photo editing tool (\eg, Lightroom), we export the editing parameters as a preset in the filters/LUTs format to process a set of images automatically. 
Meanwhile, we transfer the color style of the retouched image to other images via Neural Preset. 
As shown in Fig.\,\ref{fig:filterlutcomp}, filters/LUTs fail to convert images with diverse color styles to a consistent color style (Fig.\,\ref{fig:filterlutcomp}\,(b)). For example, after applying  filters/LUTs, bright images will be overexposed while dark images still remain dark. In contrast, with the color style parameters extracted from the retouched image, Neural Preset provides results with more consistent color styles (Fig.\,\ref{fig:filterlutcomp}\,(c)). Refer to Appendix\,\ref{appendix:as_presets} for more results of applying the same color style to different input images via Neural Preset.

\begin{table}[t]
  \begin{center}
\setlength{\tabcolsep}{4pt}
\small
\begin{tabular}{cc|c}
\toprule
\multicolumn{3}{c}{At 8K Resolution (7680 $\times$ 4320)}\\
\midrule
Patch Size & Total Patches & GPU Inference Time $\downarrow$ / Memory $\downarrow$ \\
\midrule
256 & 507 & 0.1026\,s / \textbf{1.83\,GB} \\
\underline{512} & \underline{127} & \underline{0.0613\,s} / \underline{1.96\,GB} \\
1024 & 32 & 0.0590\,s / 2.14\,GB \\
2048 & 8 & 0.0582\,s / 2.79\,GB \\
4096 & 2 & 0.0575\,s / 5.38\,GB \\
8092 & 1 & \textbf{0.0569\,s} / 8.77\,GB \\
\bottomrule
\end{tabular}
    \caption{\textbf{Effects of Different DNCM Input Patch Sizes.}
    We validate the inference time and memory cost of Neural Preset with different DNCM input patch sizes on 8K images. In Table\,\ref{tab:speed_size}, we use a patch size of 512 (the underlined row).}\label{tab:our_speed}
\vspace{-0.5cm}
\end{center}
\end{table}

\subsection{Ablation Studies}\label{sec:ablation}

\noindent\textbf{Input Patch Size for DNCM.}\quad
The input patch size used for DNCM can affect the inference time and memory footprint of Neural Preset. 
Intuitively, processing a large number of pixels in parallel (\ie, using a larger patch size) will give a lower inference time but at the cost of a higher memory consumption.
From Table\,\ref{tab:our_speed}, setting a patch size $> 512$ brings a limited speed improvement on a Nvidia RTX3090 GPU but a significant increase in memory usage. This is because the 512 patch size already uses up all GPU cores.

\medskip
\noindent\textbf{DNCM \vs \;CNN Color Mapping.}\quad We experiment with constructing the proposed two-stage color style transfer pipeline using two CNNs, \eg, two autoencoders~\cite{UNet}, with all other settings unchanged.
The results in Fig.\,\ref{fig:ablation_ncm_cnn} visualize our analysis in 
Appendix\,\ref{appendix:optimization}:
using DNCM instead of CNN for color mapping prevents our self-supervised training strategy from converging to a trivial solution.

\begin{figure}[t]
\centering
\includegraphics[width=0.99\linewidth]{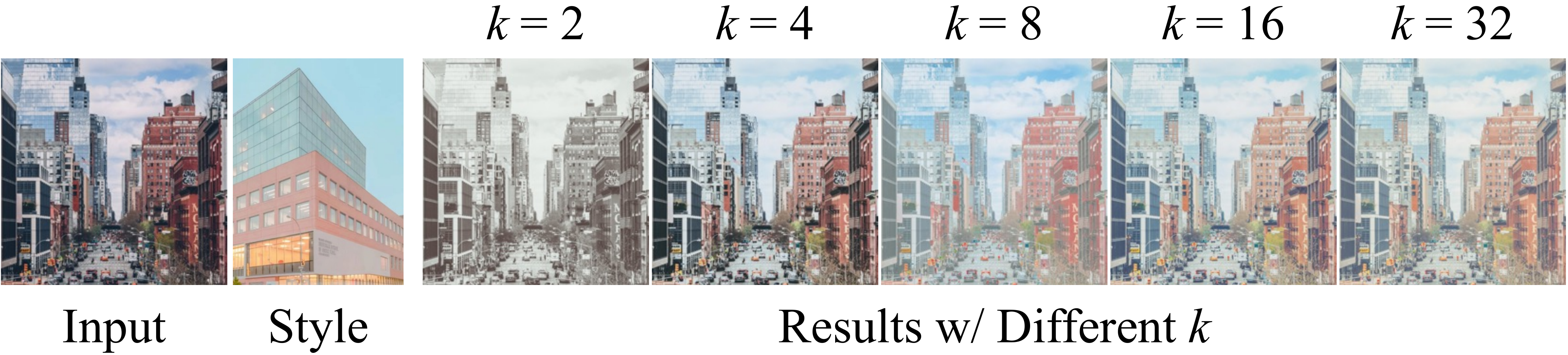}
{\begin{center}
\vspace{-0.3cm}
\caption{\textbf{Results of Neural Preset with Different Values of $k$.} These visual results are consistent with those in Table\,\ref{tab:ablation_k}.
}
\label{fig:ablation_k}
\end{center}
}
\vspace{-0.8cm}
\end{figure}

\medskip
\noindent\textbf{Parameter Dimension $k$ in DNCM.}\quad 
Table\,\ref{tab:ablation_k} shows that setting $k < 16$ significantly decreases style similarity but has small effect on content similarity, while setting $k > 16$ leads to better performances but also requires longer inference time.
Hence, we use $k = 16$, which provides a good balance of speed and performance.
Fig.\,\ref{fig:ablation_k} displays the results with different values of $k$, demonstrating that $k$ also has a huge impact on qualitative results.

\begin{figure}[t]
\centering
\includegraphics[width=0.99\linewidth]{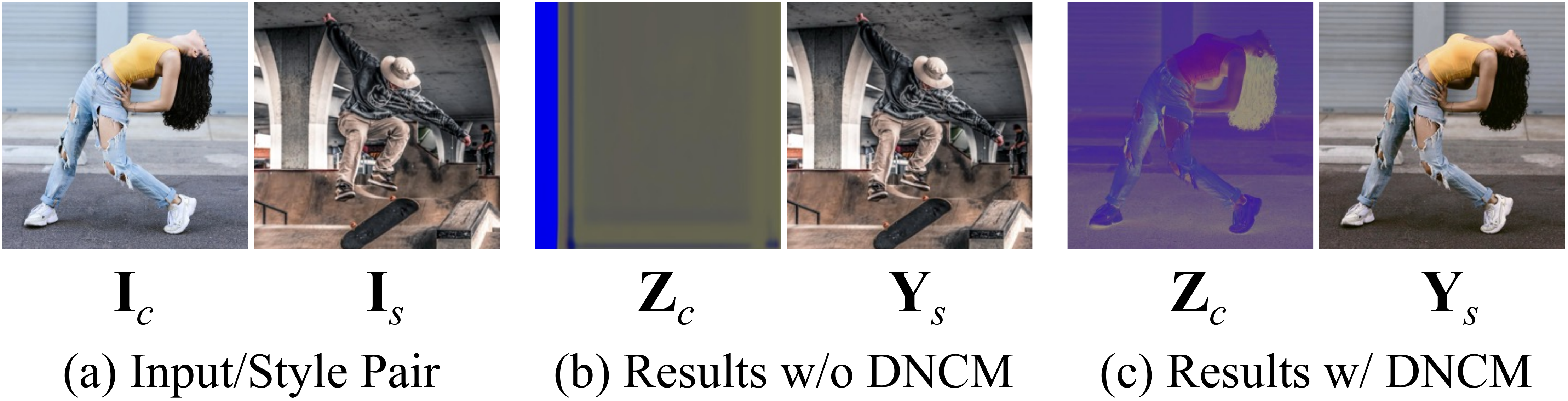}
{\begin{center}
\vspace{-0.3cm}
\caption{\textbf{Ablation of DNCM in Our Pipeline.} 
Building our two-stage pipeline via two CNNs, \eg, two autoencoders~\cite{UNet}, causes our
SSL training to converge to a trivial solution, where the first stage can be any function, and the second stage is an identity function \wrt the style image (see (b)). Applying DNCM can help avoid such a trivial solution (see (c)). 
Symbols are from Fig.\,\ref{fig:pipeline}.
}
\label{fig:ablation_ncm_cnn}
\end{center}
}
\vspace{-0.3cm}
\end{figure}

\begin{table}[t]
  \begin{center}
\setlength{\tabcolsep}{5pt}
\small
\begin{tabular}{c|cccccc}
\toprule
$k$ & 2 & 4 & 8 & \underline{16} & 32 \\
\midrule
\;\;\;\;Style Similarity $\uparrow$ & 0.128 & 0.510 & 0.636 & \underline{0.746} & \textbf{0.769} \\
Content Similarity $\uparrow$ & 0.765 & \textbf{0.823} & 0.781 & \underline{0.771} & 0.764 \\
\bottomrule
\end{tabular}
    \caption{\textbf{Results of Neural Preset with Different Values of $k$.} 
    The hyper-parameter $k$ has a huge impact on style similarity.
    Other figures and tables are based on $k=16$ (the underlined column).
    }\label{tab:ablation_k}
\vspace{-0.3cm}
\end{center}
\end{table}

 \begin{figure}[t]
\centering
\includegraphics[width=0.99\linewidth]{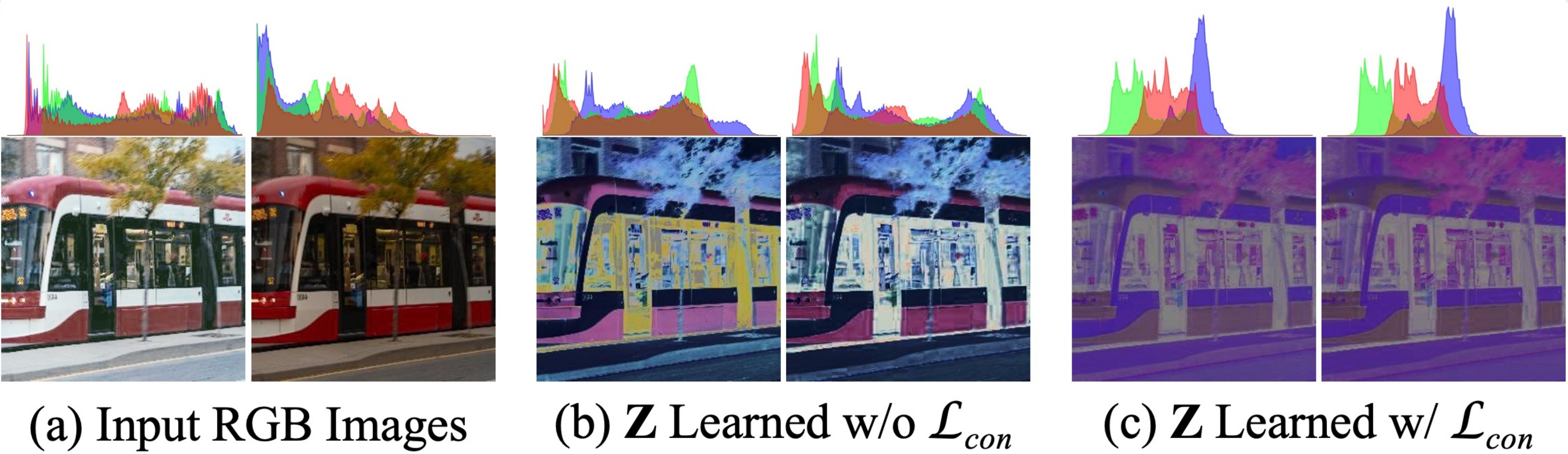}
{\begin{center}
\vspace{-0.3cm}
\caption{\textbf{Impact of $\mathcal{L}_{con}$.}
We show the RGB color histogram of each image at the top.
Applying $\mathcal{L}_{con}$ can provide a more consistent $\mathbf{Z}$, \ie, a better representation of the ``image content'', for two images with the same content but different color styles.
}
\label{fig:ablation_Lcon}
\end{center}
}
\vspace{-0.7cm}
\end{figure}

\medskip
\noindent\textbf{Effectiveness of $\mathcal{L}_{con}$.}\quad
We visualize the ``image content'' output by the first stage of our pipeline in Fig.\,\ref{fig:ablation_Lcon}. Applying $\mathcal{L}_{con}$ provides a more consistent $\mathbf{Z}$ to represent the ``image content''. Besides, our experiments also show that learning Neural Preset with $\mathcal{L}_{con}$ can yield better results, \ie, $\mathcal{L}_{con}$ helps the pipeline converge better.

 \begin{figure}[t]
\centering
\includegraphics[width=0.99\linewidth]{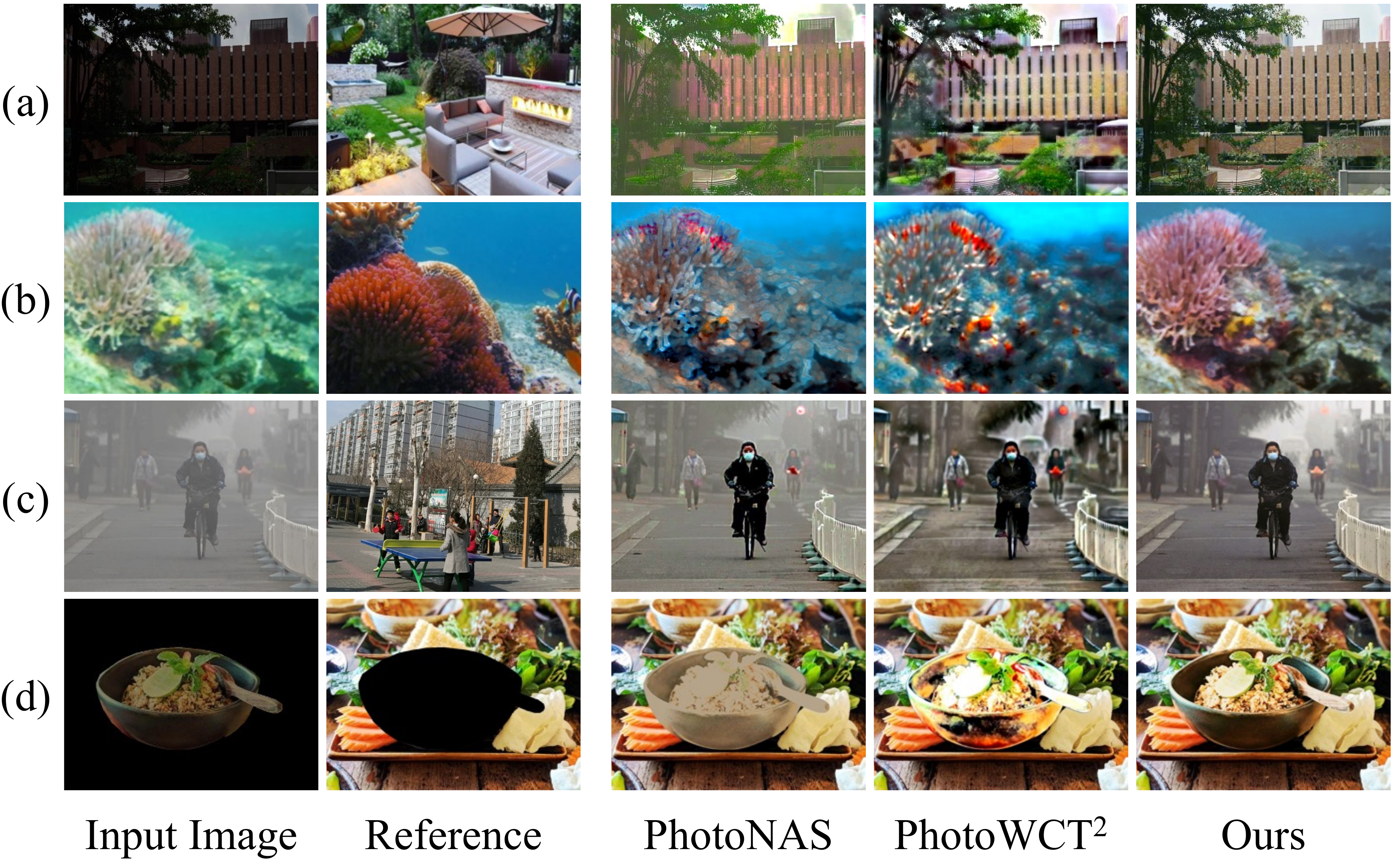}
{\begin{center}
\vspace{-0.3cm}
\caption{\textbf{Applied to Other Color Mapping Tasks.} (a) Low-light image enhancement; (b) underwater image correction; (c) image dehazing; (d) image harmonization (the input image is the foreground while the reference image is the background).
}
\label{fig:app_other_tasks}
\end{center}
}
\vspace{-0.5cm}
\end{figure}

 \begin{figure}[t]
\centering
\includegraphics[width=0.99\linewidth]{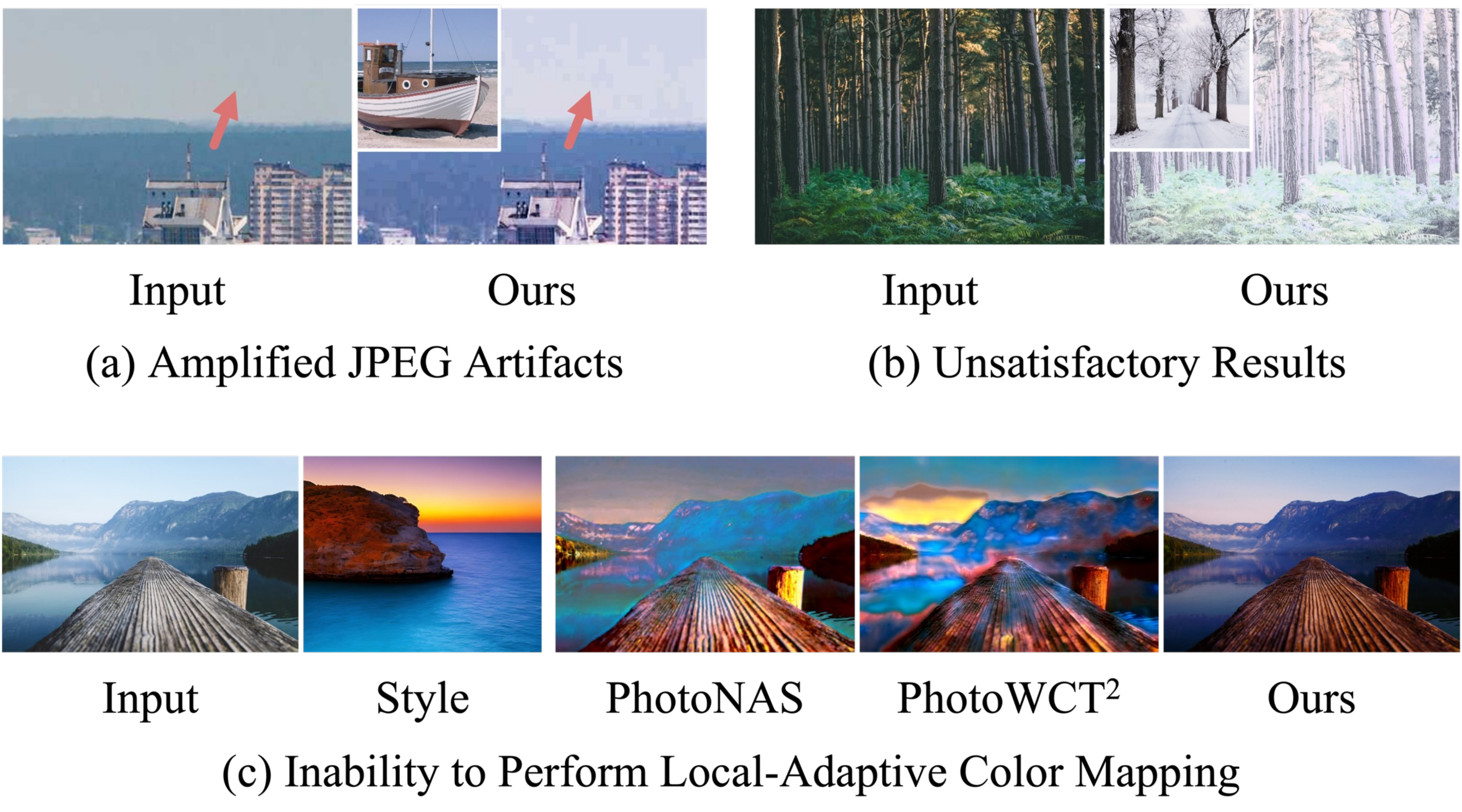}
{\begin{center}
\vspace{-0.3cm}
\caption{\textbf{Limitations of Neural Preset.}
For (a) and (b), the top-left corner shows the style image. (a) JPEG artifacts may be amplified (see red arrows) if the input image has a high compression ratio. (b) Results may be unsatisfactory if some colors in the input image (\eg, green color) do not exist in the style image. (c) Our method fails to map blue sky/water in an image to different colors separately. Although prior methods may handle blue sky/water separately, they typically cause heavy artifacts.
}
\label{fig:limitations}
\end{center}
}
\vspace{-0.7cm}
\end{figure}

\section{Applications}\label{sec:applications}

With the proposed self-supervised strategy, Neural Preset is able to learn general knowledge of color style transfer from large-scale data. As a result, given a reference image, our trained model can be naturally applied to other color mapping tasks without fine-tuning. We evaluate four tasks: low-light image enhancement~\cite{low-light-survey}, underwater image correction~\cite{underwater-survey}, image dehazing~\cite{dehazing-survey}, and image harmonization~\cite{harmonization-survey}. These tasks are challenging for universal color style transfer models as the input images typically come from highly degraded domains. 
The qualitative results in Fig.\,\ref{fig:app_other_tasks} show that Neural Preset significantly outperforms prior color style transfer methods on all four tasks. 
We notice that our DNCM effectively avoids heavy visual artifacts produced by other methods, and our color normalization stage successfully decouples color styles from degraded input images. Refer to 
Appendix\,\ref{appendix:apply_other_task} 
for comparisons with more color style transfer methods on these four tasks.

Besides, training DNCM (Fig.\,\ref{fig:ncm}) for color mapping tasks using pairwise data is straightforward. We experiment with DNCM on the pairwise datasets of image harmonization~\cite{DIH,cvpr2022Cong,Harmonizer} and image color enhancement~\cite{hdrnet,zeng2020lut,Exposure}. 
Without whistles and bells, we get top-level performances on both tasks. Refer to 
Appendix\,\ref{appendix:ncm_application} 
for details and results.

On-device~\cite{on_device1} (\eg, mobile or browser) deployment is also vital in practice. It expects a small computational overhead in the client to support real-time UI responses and a small amount of data exchanges between the client/server to save network bandwidth, which are hard to achieve by existing color style transfer models. Instead, Neural Preset enables distributed deployment, which is friendly for on-device applications. Refer to 
Appendix\,\ref{appendix:distributed} 
for details.

\section{Conclusion}
In this paper, we have presented a simple but effective Neural Preset technique for color style transfer. Benefited by the proposed DNCM and two-stage pipeline, Neural Preset has shown significant improvements over existing state-of-the-art methods in various aspects.
In addition, we have also explored several applications of our method. For example, directly applying Neural Preset to other color mapping tasks without fine-tuning.

Nonetheless, Neural Preset does have limitations. First, if the input is compressed by JPEG with a high compression ratio, the existing JPEG artifacts may be amplified in the output (Fig.\ref{fig:limitations}\,(a)).
Second, it may fail to transfer color styles between images with very different inherent colors (Fig.\ref{fig:limitations}\,(b)).
Third, it cannot perform local-adaptive color mapping to transfer the same color in an image to different colors (Fig.\ref{fig:limitations}\,(c)).
A possible future work is to address these limitations. For example, developing auxiliary regularization to alleviate the effects of JPEG artifacts or incorporating appropriate user interactions for complex cases that involve changing image inherent colors.



\appendix
\section*{Appendices}
\addcontentsline{toc}{section}{Appendices}
\renewcommand{\thesubsection}{\Alph{subsection}}

\subsection{Optimization Problem of Neural Preset}\label{appendix:optimization}

Here we analyze (1) how to derive our training constraints from the fundamental color style transfer objective and (2) why performing color mapping via the proposed DNCM is necessary for our training strategy.

Consider the objective of color style transfer. Given an input image $\mathbf{I}_{c}$ and a style image $\mathbf{I}_{s}$ that have different content and color styles, we aim to learn a model $H$ to transfer the color style of $\mathbf{I}_{s}$ to $\mathbf{I}_{c}$ by minimizing the objective: 
\begin{equation}\label{eq:basic_objective}
    \min \; \mathbb{E}_{\mathbf{I}_{c},\mathbf{I}_{s},\mathbf{G}_{s} \sim p_{\mathbf{I}}} \big[\, | \mathbf{G}_{s} - H (\mathbf{I}_{c}, \mathbf{I}_{s}) | \,\big],
\end{equation}
where $p_{\mathbf{I}}$ represents the distribution of images, and $\mathbf{G}_{s}$ is the ground truth stylized image that has the same image content as $\mathbf{I}_{c}$ and the consistent color style as $\mathbf{I}_{s}$.

The idea of our two-stage pipeline is to divide $H$ into two sub-functions, \ie, two stages, to remove the original image color style of $\mathbf{I}_{c}$ before applying a new one, as:
\begin{equation}\label{eq:our_objective}
    \min \; \mathbb{E}_{\mathbf{I}_{c},\mathbf{I}_{s},\mathbf{G}_{s} \sim p_{\mathbf{I}}} \big[\, | \mathbf{G}_{s} - S_{2} ( S_{1} (\mathbf{I}_{c}), \mathbf{I}_{s}) | \,\big],
\end{equation}
where $S_1$ and $S_2$ denote the sub-functions corresponding to the two stages of our pipeline.
As $\mathbf{G}_{s}$ is typically unavailable in practice, we generate pseudo input and style images to approximate the above optimization problem.
Specifically, we add random color perturbations (\eg, LUTs or filters) to each image $\mathbf{I}$ to create a set of $n$ images $\{\mathbf{I}_{1}, \dots, \mathbf{I}_{n}\}$ with the same content but different color styles. 
We denote indexes by $i, j \in \{0, \dots , n\}$ in the following context.
Thus, Eq.\,\ref{eq:our_objective} can be approximated with perturbed images, as:
\begin{equation}\label{eq:our_objective_approx}
    \min \; \mathbb{E}_{\mathbf{I} \sim p_{\mathbf{I}}} \Big[\, \sum_{i}^{n} \sum_{j}^{n} \Big| \mathbf{I}_{j} - S_{2} ( S_{1} (\mathbf{I}_{i}), \mathbf{I}_{j}) \Big| \,\Big].
\end{equation}
Given any fixed $S_1$, the optimal $S_2^{*}$ should satisfy: 
\begin{equation}\label{eq:optimal_S2}
    \mathbf{I}_{j} =
        S_2^{*} (S_{1} (\mathbf{I}_{j}), \mathbf{I}_{j}) = S_2^{*} (S_{1} (\mathbf{I}_{i}), \mathbf{I}_{j}).
\end{equation}
Eq.\,\ref{eq:optimal_S2} reveals a possible optimization issue of Eq.\,\ref{eq:our_objective_approx} --
if we model $S_1$ and $S_2$ by end-to-end CNNs like autoencoders~\cite{UNet}, optimizing only Eq.\,\ref{eq:our_objective_approx} via gradient-based algorithms can easily make $S_1$ and $S_2$ converge to a trivial solution to satisfy Eq.\,\ref{eq:optimal_S2}: $S_2^{*}$ becomes an identity function \wrt $\mathbf{I}_{j}$, while $S_1$ can be any function.
Formally, for a real input and style image pair $\mathbf{I}_c$ and $\mathbf{I}_s$ with different content, the possible trivial solution is:
\begin{equation}\label{eq:bad_S2}
        \mathbf{I}_{s} = S_2^* (\phi, \mathbf{I}_{s}), \quad \phi := S_1(\mathbf{I}_{c}),
\end{equation}
where $S_2^*$ always output $\mathbf{I}_{s}$ directly and ignore another input $\phi$. Such a solution is undesired because the color style transfer objective (Eq.\,\ref{eq:basic_objective}) requires the output image have the same content with $\mathbf{I}_{c}$.

To overcome the above problem, modeling $S_1$ and $S_2$ via DNCM instead of end-to-end CNNs is necessary. 
DNCM inputs the color style parameters $E(\mathbf{\tilde{I}}_{j})$ rather than the image $\mathbf{I}_{j}$. Since the dimensions of $E(\mathbf{\tilde{I}}_{j})$ is much lower than $\mathbf{I}_{j}$, it prevents $S_2$ from being an identity function \wrt $\mathbf{I}_{j}$ and forces $S_1$ to be involved in computing the optimal solution. Specifically, we define: 
\begin{equation}\label{eq:S1_S2}
\begin{split}
    S_1 (\mathbf{I}_{i}) &:= \textit{nDNCM}(\mathbf{I}_{i}, E(\mathbf{\tilde{I}}_{i})), \\
    S_2 (S_1(\mathbf{I}_{i}), \mathbf{I}_{j}) &:= \textit{sDNCM}(S_1(\mathbf{I}_{i}), E(\mathbf{\tilde{I}}_{j}))).
\end{split}
\end{equation}
The formulas/symbols in Eq.\,\ref{eq:S1_S2} are equivalent to those in Sec.\,{\color{linkCol} 3.2} of the paper. Benefited by DNCM, optimizing only Eq.\,\ref{eq:our_objective_approx} is sufficient for preventing $S_1$ and $S_2$ from converging to the trivial solution shown in Eq.\,\ref{eq:bad_S2}.  
There are other possible approaches to avoid such a trivial solution without using DNCM, \eg, modeling the optimization of our pipeline as a bi-level optimization~\cite{bilevel-survey,bilevel-grad} problem. 

Let us review Eq.\,\ref{eq:optimal_S2} by substituting in Eq.\,\ref{eq:S1_S2}, it is obvious that constraining $S_1(\mathbf{I}_i)$ and $S_1(\mathbf{I}_j)$ to be consistent will make $S_2^*$ easier to obtain. Hence, we interpret Eq.\,\ref{eq:our_objective_approx} as a constrained optimization problem:
\begin{equation}\label{eq:constrained_objective}
    \begin{split}
         \min \; \mathbb{E}_{\mathbf{I} \sim p_{\mathbf{I}}} \Big[\, \sum_{i}^{n} \sum_{j}^{n} \Big| \mathbf{I}_{j} - S_{2} ( S_{1} (\mathbf{I}_{i}), \mathbf{I}_{j}) \Big| \,\Big] \\
            s.t. \quad  \sum_{i}^{n} \sum_{j}^{n} \Big| S_{1} (\mathbf{I}_{j}) - S_{1} (\mathbf{I}_{i}) \Big|  = 0 .
    \end{split}
\end{equation}
By using Penalty or Augmented Largangian~\cite{Hestenes1969MultiplierAG} methods, Eq.\,\ref{eq:constrained_objective} can be reformulated as an unconstrained optimization problem. For example, the Penalty method produces:
\begin{equation}\label{eq:unconstrained_objective}
\begin{split}
   & \min \; \mathbb{E}_{\mathbf{I} \sim p_{\mathbf{I}}} [\, 
   \mathcal{L}_{rec} + \lambda \, \mathcal{L}_{con} \,], \\ 
   & \mathcal{L}_{rec} := \sum_{i}^{n} \sum_{j}^{n} \Big| \mathbf{I}_{j} - S_{2} ( S_{1} (\mathbf{I}_{i}), \mathbf{I}_{j}) \Big|, \\ 
   & \mathcal{L}_{con} := \sum_{i}^{n} \sum_{j}^{n} \Big| S_{1} (\mathbf{I}_{j}) - S_{1} (\mathbf{I}_{i}) \Big|,
\end{split}
\end{equation}
where $\lambda$ is a penalty coefficient. 
If we set $n=2$, this new optimization objective is formally equivalent to our training constraint defined in Sec.\,{\color{linkCol} 3.3} of the paper.

\subsection{Details of Our Improved Quantitative Metrics}\label{appendix:quantitive_eval}

Below we describe the implementation details of our improved quantitative metrics for color style transfer.

\begin{figure}[t]
\centering
\includegraphics[width=0.99\linewidth]{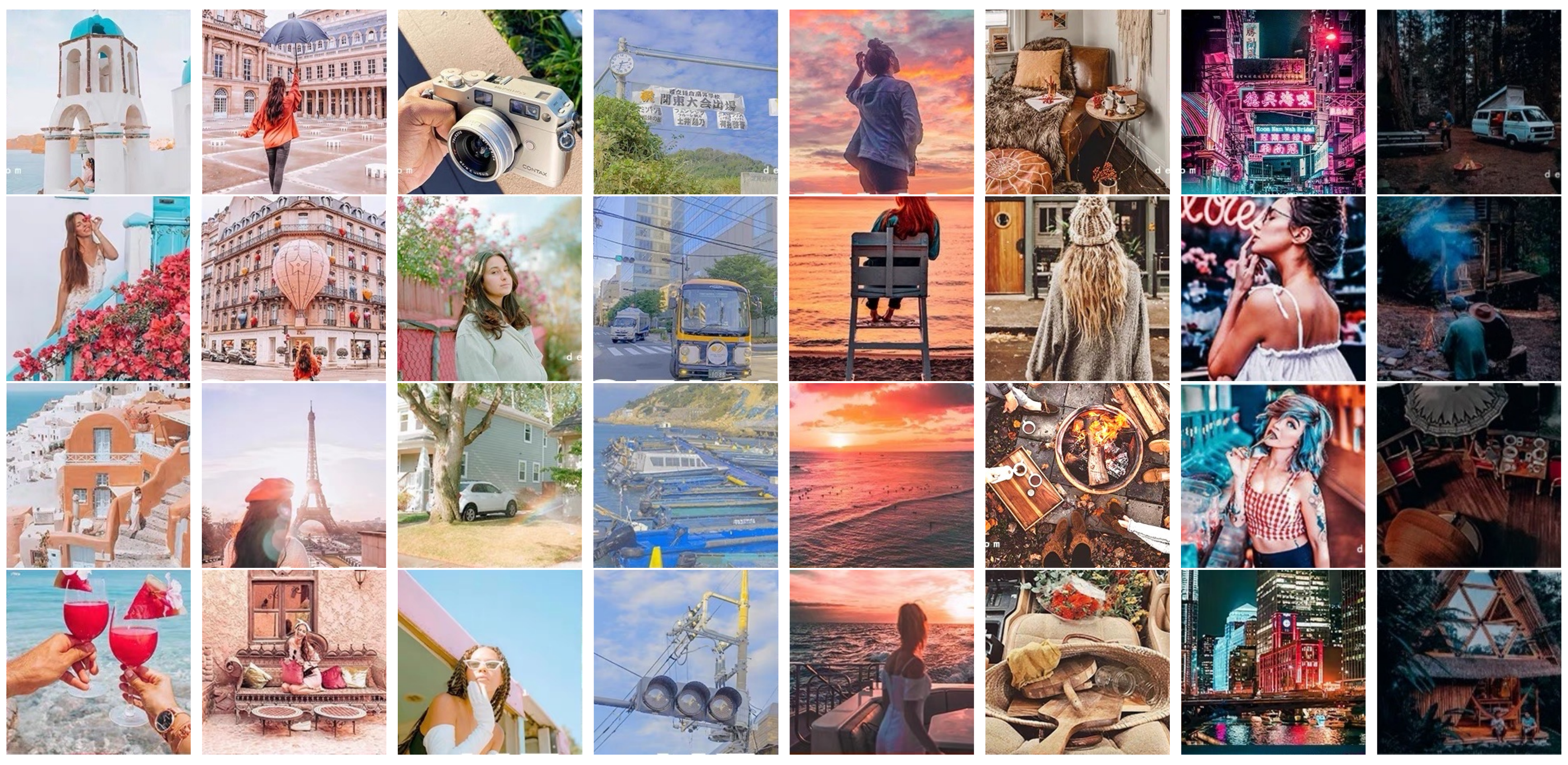}
{\begin{center}
\vspace{-0.3cm}
\caption{\textbf{Data Used to Train Our Color Style Discriminator.} We display some samples from the dataset. The four samples in each column belong to the same color style category.}
\label{fig:style_dis_dataset}
\end{center}
}
\vspace{-0.5cm}
\end{figure}

\medskip
\noindent\textbf{Style Similarity Metric.}\quad 
We first build a dataset consists of 700+ color style categories, each containing 6-10 images with the same color style retouched by human experts (see Fig.\,\ref{fig:style_dis_dataset}).
We then train a discriminator model $D$ on this dataset as our style similarity metric. 
Specifically, for an image $\mathbf{I}$ in the dataset, we use $\mathbf{I}_{p}$ and $\mathbf{I}_{n}$ to denote a positive sample from the same category as $\mathbf{I}$ and a negative sample from any other category, respectively.
We optimize $D$ to distinguish different color styles via minimizing the following loss from LS-GAN~\cite{LSGAN}:
\begin{equation}\label{eq:Ldis}
    \mathcal{L}_{dis} = || D(\mathbf{I}, \, \mathbf{I}_{p}) - 1 ||_{2} + || D(\mathbf{I}, \, \mathbf{I}_{n}) - 0 ||_{2}.
\end{equation}
The trained $D$ will output a score between $[0, 1]$, which represents the style similarity between two input images. 
The score tends to be $1$ if the two input images have similar color styles and $0$ otherwise. 
The architecture of $D$ is adapted from~\cite{pix2pix2017}. We use the Adam~\cite{Adam} optimizer to train $D$ for 120 epochs. The initial learning rate is $1e^{-4}$ and is multiplied by $0.1$ after every 50 epochs.

 \begin{figure}[t]
\centering
\includegraphics[width=0.99\linewidth]{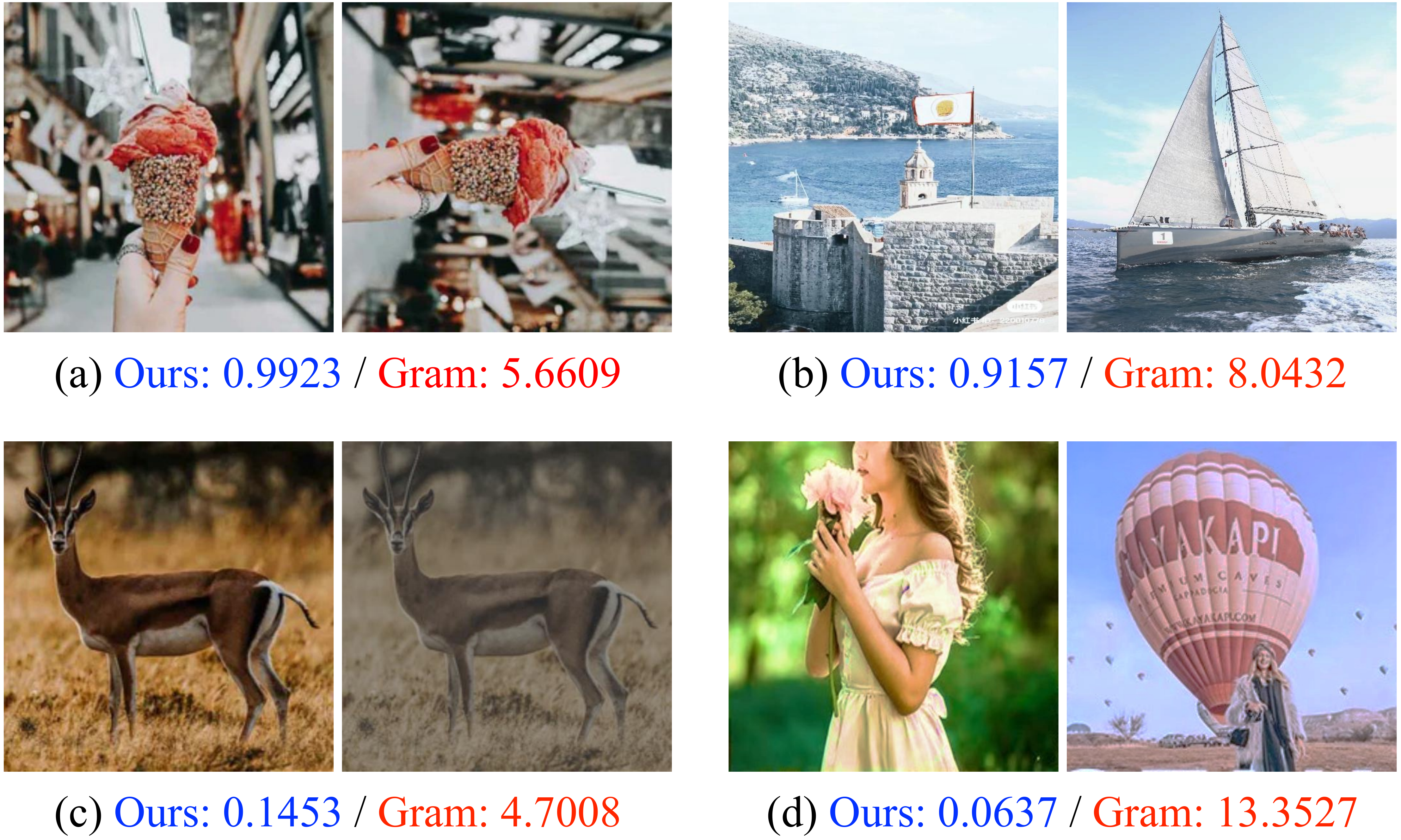}  
{\begin{center}
\vspace{-0.3cm}
\caption{\textbf{Predicted Style Similarity.} The blue number below each image pair is the style similarity (between $[0, 1]$) predicted by our metric (\ie, the color style discriminator), and a higher value means that the two images have more similar color styles. 
The red number below each image pair is the style similarity (larger than 0) predicted by the Gram metric, and a lower value means that the two images have more similar color styles.
We compare two metrics in four cases: (a) two images with the same content and color style; (b) two images with different content but similar color styles; (c) two images with the same content but different color styles; (d) two image with different content and color styles. 
}
\label{fig:style_dis_result}
\end{center}
}
\vspace{-0.7cm}
\end{figure}

 \begin{figure}[t]
\centering
\includegraphics[width=0.99\linewidth]{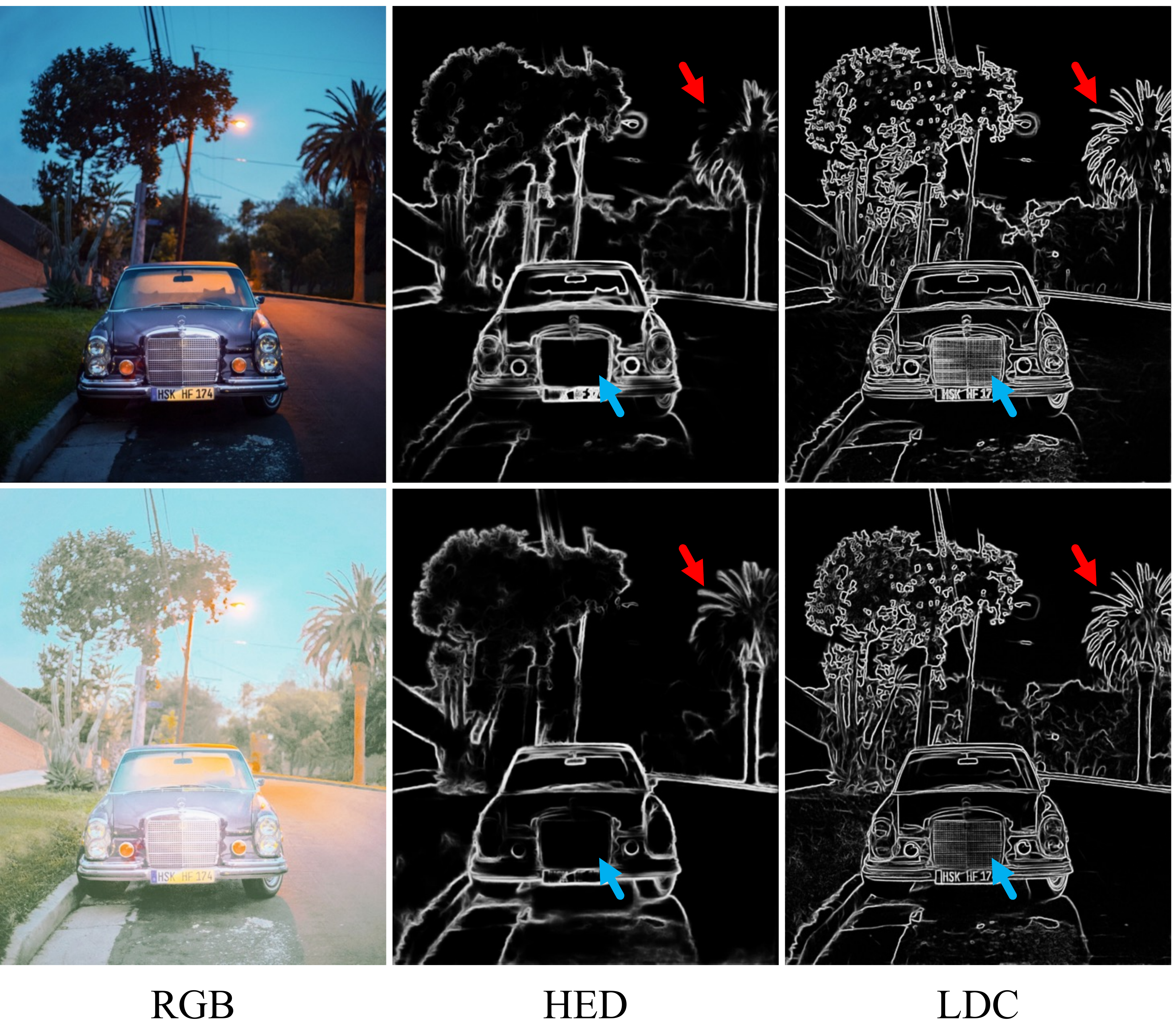}
{\begin{center}
\vspace{-0.3cm}
\caption{\textbf{Predicted Edges for Content Similarity Calculation.}
For a more precise quantitative evaluation of content similarity, we replace HED~\cite{HN_Edge} (used by prior works~\cite{PhotoWCT, WCT2,an2019ultrafast}) with LDC~\cite{xsoria2022ldc}. 
LDC can provide finer (see blue arrows) and more accurate edges (see red arrows) than HED.
}
\label{fig:edge_comp}
\end{center}
}
\vspace{-0.7cm}
\end{figure}

Our metric, \ie, the trained $D$ focuses more on comparing image color styles and ignores other irrelevant information, such as the photorealism of the image content.
Besides, $D$ predicts the style similarity score based on not only on the statistic of color values but also on image properties, \eg, whether the two images have the same global contrast. 
To demonstrate that our newly proposed metric is more meaningful than the Gram metric used by prior works, we compare the style similarity predicted by our metric and the previously used Gram metric in Fig.\,\ref{fig:style_dis_result}. 
Note that a lower Gram value means a higher similarity. Gram $<$ 5 usually indicates very similar, while Gram $>$ 8 indicates a huge difference. 
We can see that both metrics work well if two images have exactly the same (Fig.\,\ref{fig:style_dis_result}\,(a)) or very different (Fig.\,\ref{fig:style_dis_result}\,(d)) content and color style. However, for two images with the same content but different color styles, the Gram metric may consider they have similar color styles (Fig.\,\ref{fig:style_dis_result}\,(c)). Meanwhile, for two images with different content but similar color styles, the Gram metric may consider they are different in color style (Fig.\,\ref{fig:style_dis_result}\,(b)). 
Instead, our metric gives reasonable results in these cases.

\medskip
\noindent\textbf{Content Similarity Metric.}\quad
We test the edge detection method HED~\cite{HN_Edge} used by prior works~\cite{PhotoWCT, WCT2,an2019ultrafast}, and we observe that HED fails to detect fine edges and often predict inaccurate edges, leading to an unreliable content similarity evaluation. To alleviate this problem, we suggest replacing HED with a state-of-the-art edge detection method LDC~\cite{xsoria2022ldc} to provide more precise edges. Fig.\,\ref{fig:edge_comp} compares the visual results of HED and LDC, which shows the advantages of LDC.
When computing the metric, we set the long side of the HED/LDC input images to $2048$ to preserve image textures.
After extracting edges, we compute the SSIM metric between them as the content similarity score of the two input images.

\subsection{More Results}

\subsubsection{Visual Results of Neural Preset}\label{appendix:image_results}
\begin{figure*}[th]
\centering
\includegraphics[width=0.99\linewidth]{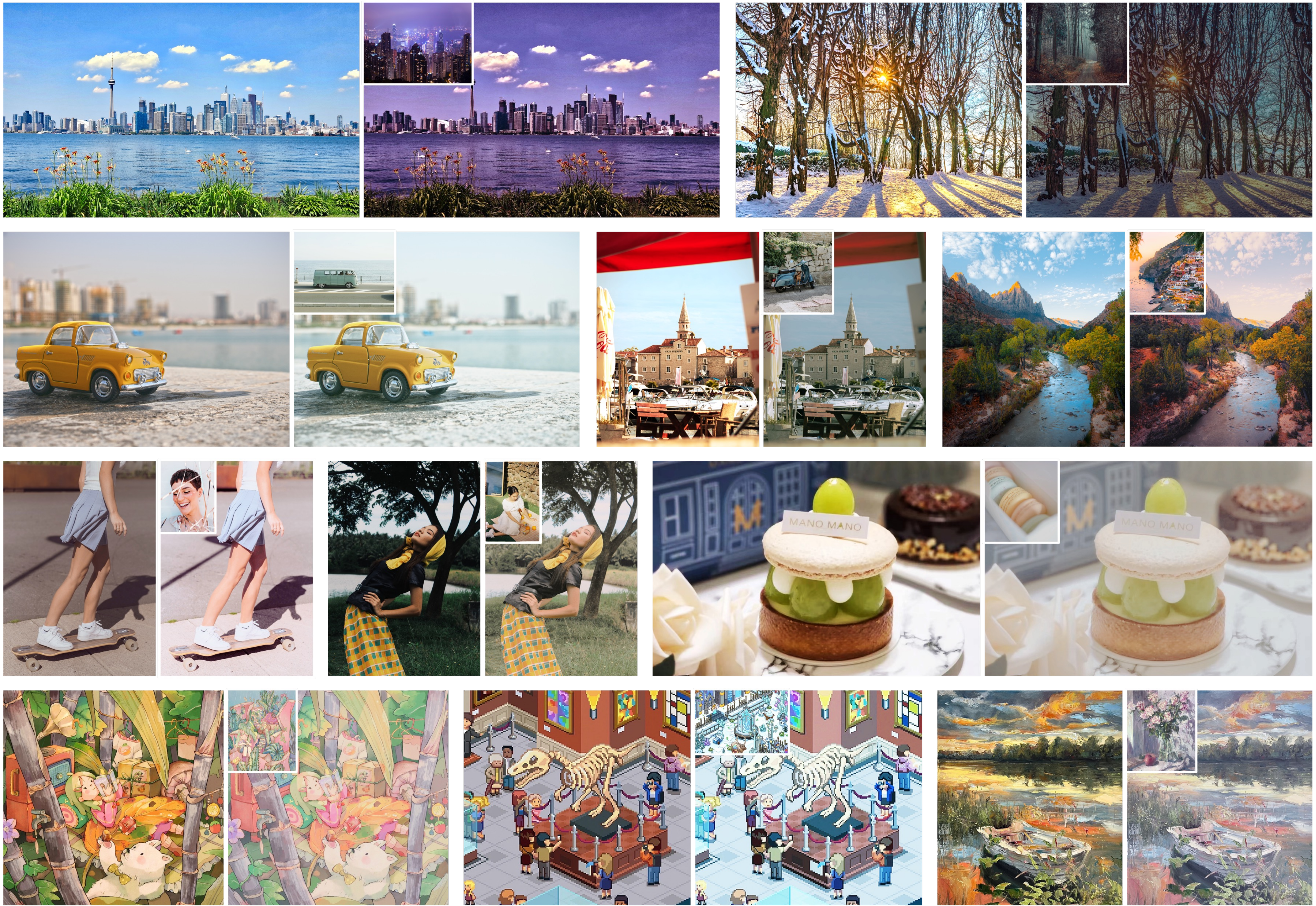}
{\begin{center}
\vspace{-0.3cm}
\caption{\textbf{Image Color Style Transfer Results of Neural Preset.} For each image pair, the left is the input image, while the right is our stylized result. The reference style image is displayed in the top-left corner. Our method is robust when generalizing to different types of images, \eg, illustrations, pixelated images, oil paintings  (see the last row).}
\label{fig:appendix_image_result}
\end{center}
}
\vspace{-0.5cm}
\end{figure*}

\begin{figure*}[th]
\centering
\includegraphics[width=0.99\linewidth]{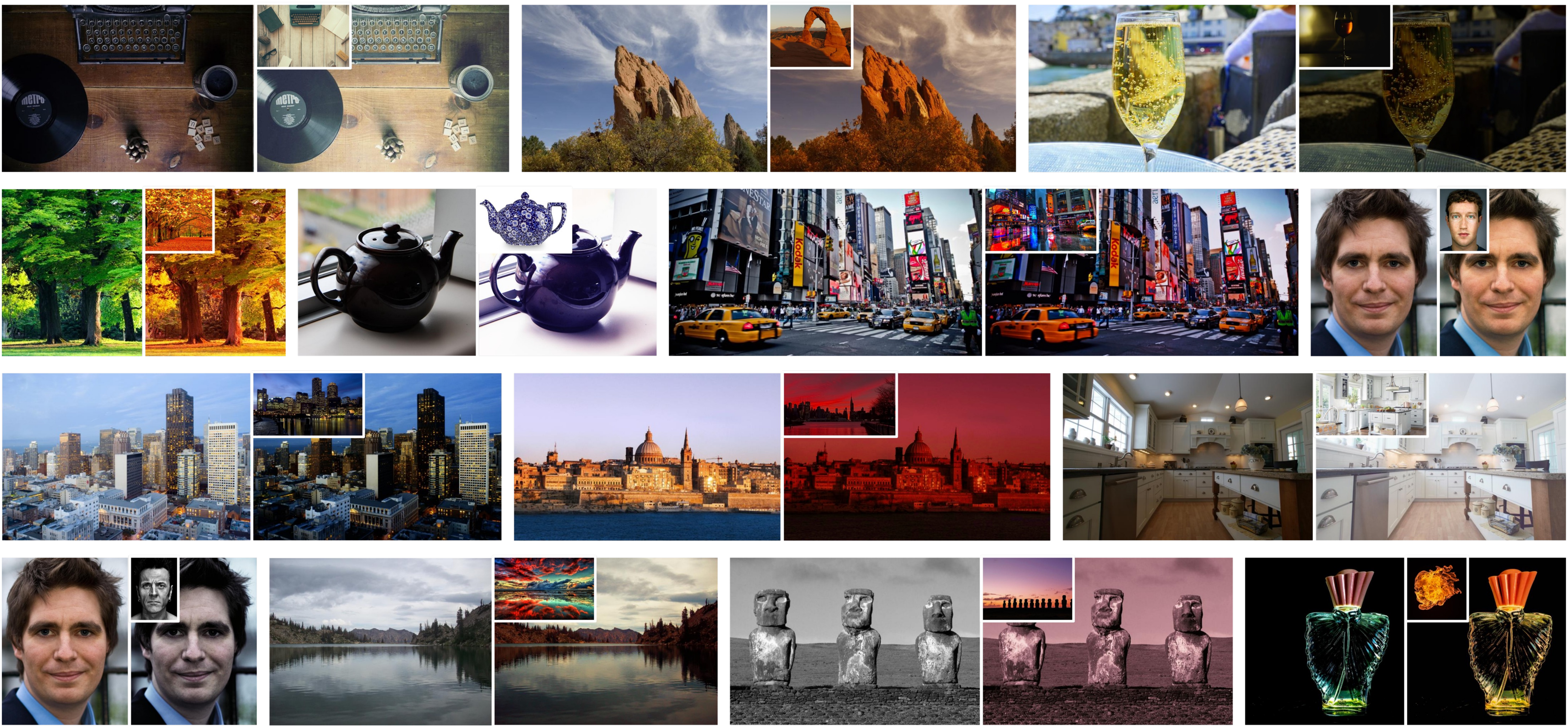}
{\begin{center}
\vspace{-0.3cm}
\caption{\textbf{Image Color Style Transfer Results of Neural Preset.} For each image pair, the left is the input image, while the right is our stylized result. The reference style image is displayed in the top-left corner. All samples we show here are from the test set provided by Luan \etal~\cite{DPST}. Neural Preset works well on most cases (see the first three rows), except for cases similar to the ones we have discussed in the limitations (see the last row).}
\label{fig:priordataset}
\end{center}
}
\vspace{-0.5cm}
\end{figure*}

\;\;\;\;\;We provide more visual results in Fig.\,\ref{fig:appendix_image_result} and Fig.\,\ref{fig:priordataset}.

\vspace{-0.3cm}
\subsubsection{Video Stylization Results of Neural Preset}\label{appendix:video_results}
\begin{figure}[t]
\centering
\includegraphics[width=0.99\linewidth]{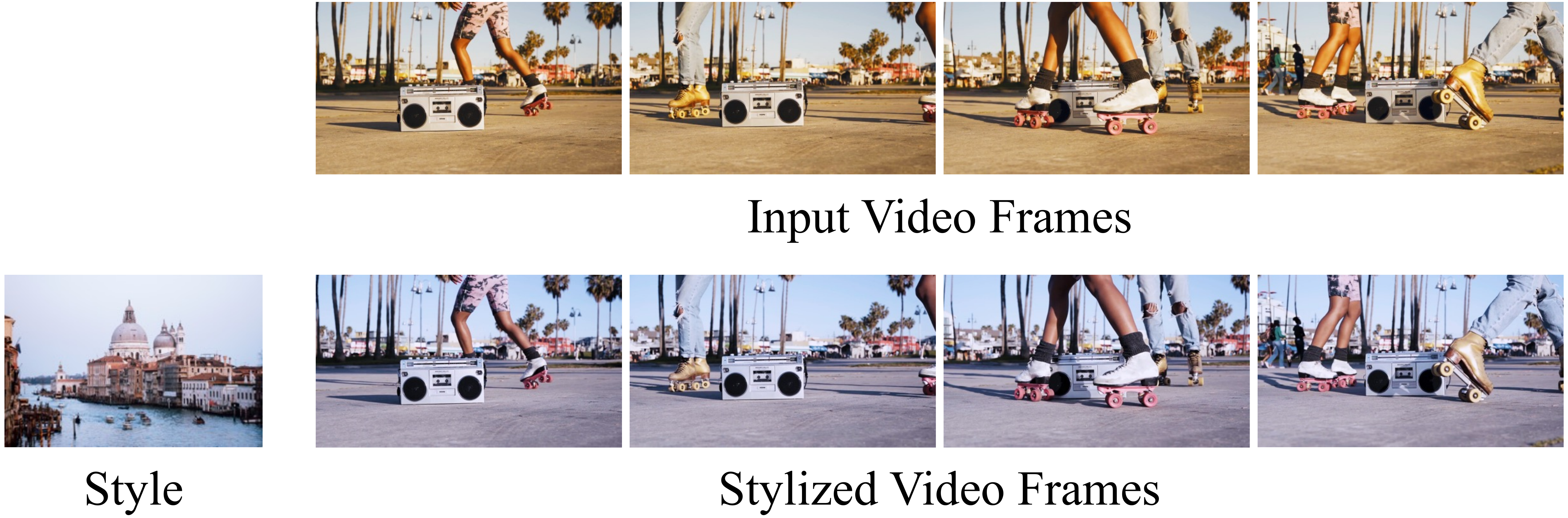}
{\begin{center}
\vspace{-0.3cm}
\caption{\textbf{Our Video Color Stylization Results.} Neural Preset provides consistent color style transfer results across video frames. 
}
\label{fig:video_result}
\end{center}
}
\vspace{-0.5cm}
\end{figure}

\;\;\;\;\;We show frames of video results in Fig.\,\ref{fig:video_result} and provide videos in our project page. 
By creating \textit{nDNCM}/\textit{sDNCM} from the first frame and using them to process subsequent frames, Neural Preset can provide consistent results across frames. 
In contrast, prior methods 
often cause flickering artifacts and post-processing like DVP~\cite{lei2022deep} should be applied.

\vspace{-0.3cm}
\subsubsection{Stylize Various Images with the Same Color Style}\label{appendix:as_presets}
\begin{figure}[t]
\centering
\includegraphics[width=0.99\linewidth]{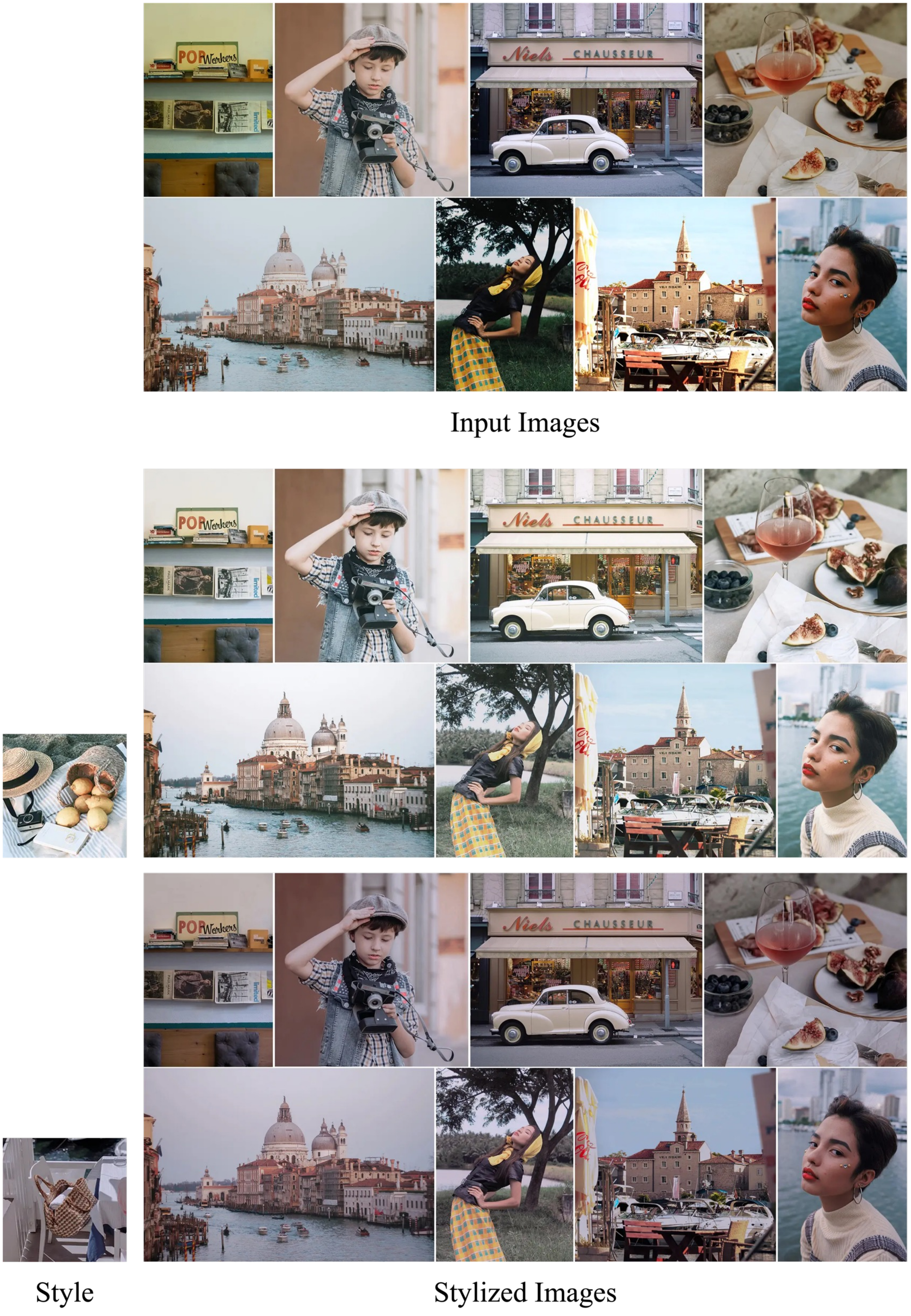}
{\begin{center}
\vspace{-0.3cm}
\caption{\textbf{Our Image Stylization Results.} Neural Preset can convert images with diverse color styles to the same color style.
}
\label{fig:as_presets}
\end{center}
}
\vspace{-0.7cm}
\end{figure}

\;\;\;\;\;Fig.\,\ref{fig:as_presets} shows the results of applying the same color style to different images through Neural Preset.


\vspace{-0.3cm}
\subsubsection{Comparison on User Study Results}\label{appendix:more-user-study_results}
\begin{figure*}[t]
\centering
\includegraphics[width=0.92\linewidth]{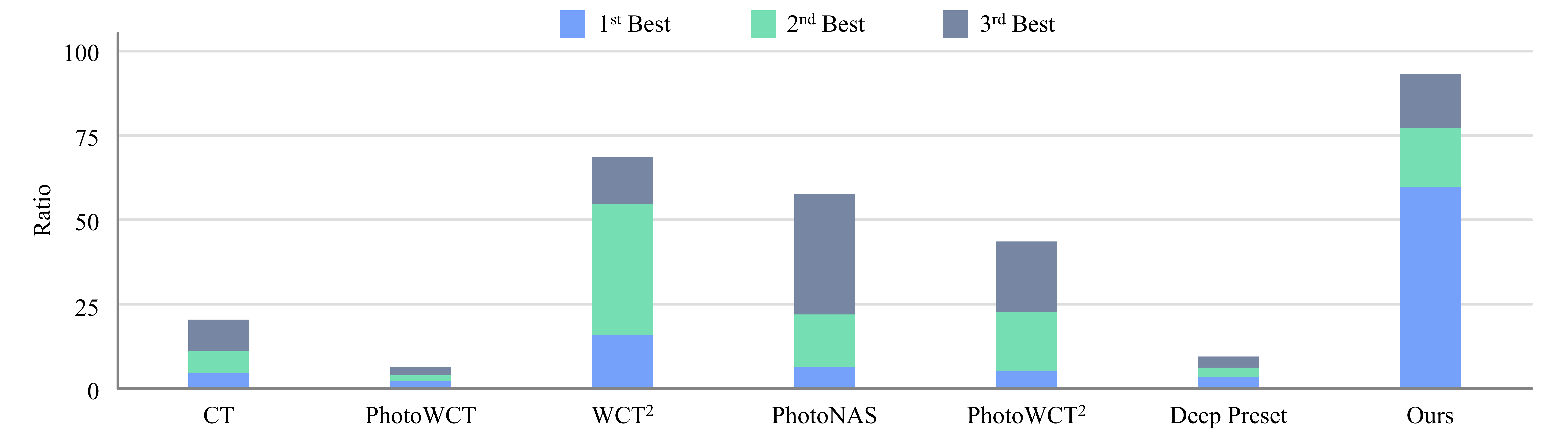}
{\begin{center}
\vspace{-0.3cm}
\caption{\textbf{Comparison on User Study Results.} 
 We display the ratios of each method being ranked as 1$^{\text{st}}$, 2$^{\text{nd}}$ and 3$^{\text{rd}}$ Best. Our Neural Preset is ranked as the Top1, Top2, and Top3 results over $61\%$, $78\%$, and $93\%$ cases, respectively.
}
\label{fig:appendix_userstudy}
\end{center}
}
\vspace{-0.5cm}
\end{figure*}

\;\;\;\;\;
In Fig.\,\ref{fig:appendix_userstudy}, we calculate the ratios of each method being ranked as 1$^{\text{st}}$ Best, 2$^{\text{nd}}$ Best, and 3$^{\text{rd}}$ Best.
Neural Preset is selected as 1$^{\text{st}}$ Best (\ie, Top1) in $61.28\%$ of cases, significantly surpassing $16.25\%$ obtained by the second-ranked WCT$^2$. In addition, Neural Preset achieves a Top3 (\ie, (1$^{\text{st}}$ + 2$^{\text{nd}}$ + 3$^{\text{rd}}$) Best) ratio of $93.02\%$.

\vspace{-0.3cm}
\subsubsection{Comparison on Applied to Other Tasks}\label{appendix:apply_other_task}
\begin{figure*}[th]
\centering
\includegraphics[width=0.99\linewidth]{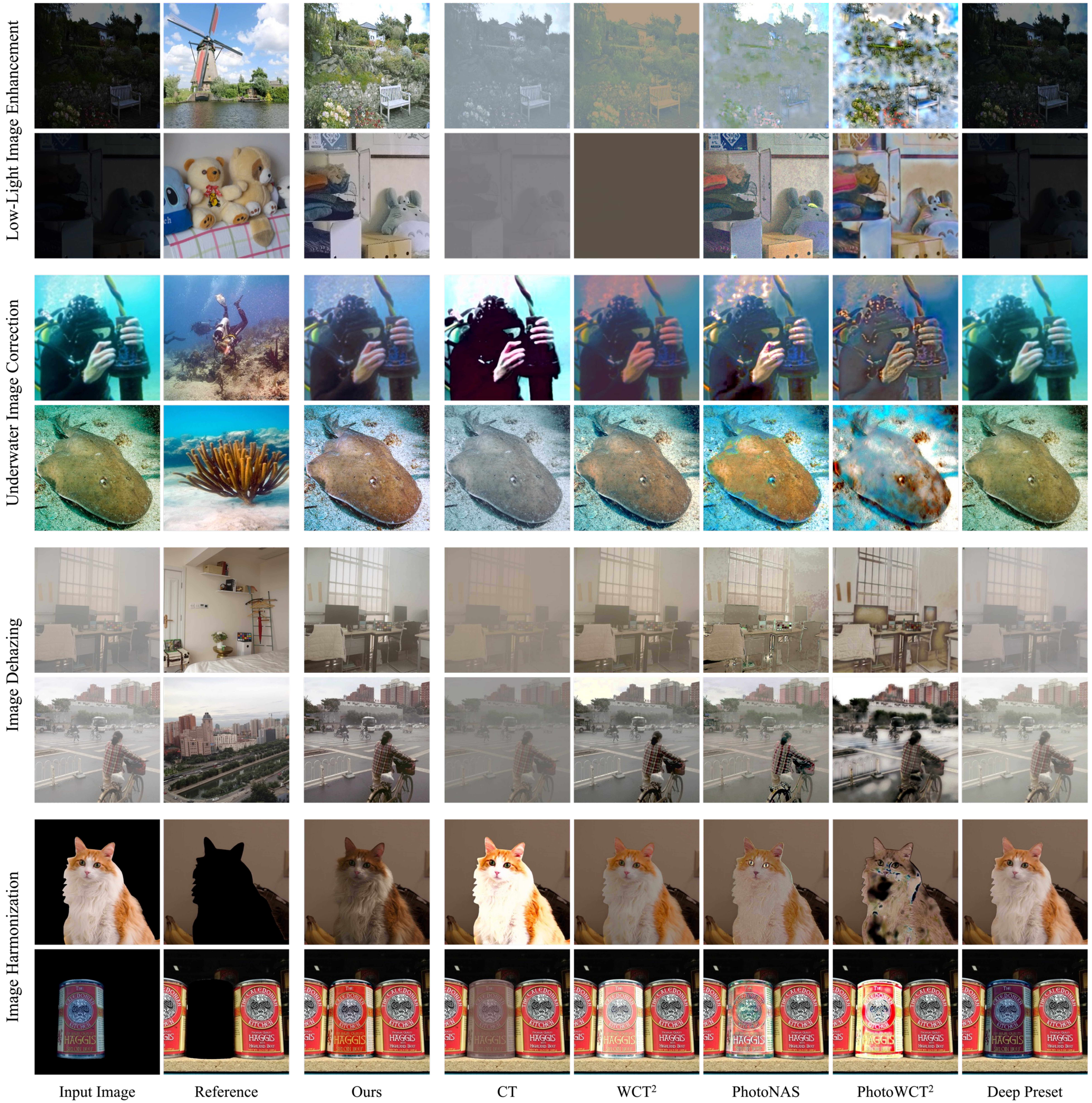}
{\begin{center}
\vspace{-0.3cm}
\caption{\textbf{Applying Color Style Transfer Methods to Other Tasks without Fine-tuning.} Our Neural Preset robustly generalizes to other color mapping tasks and surpasses previous color style transfer methods by a large margin. The datasets we used are listed in the Acknowledgments at the end of the paper.
}
\label{fig:othertask_supp}
\end{center}
}
\vspace{-0.7cm}
\end{figure*}

\begin{figure}[t]
\centering
\includegraphics[width=0.99\linewidth]{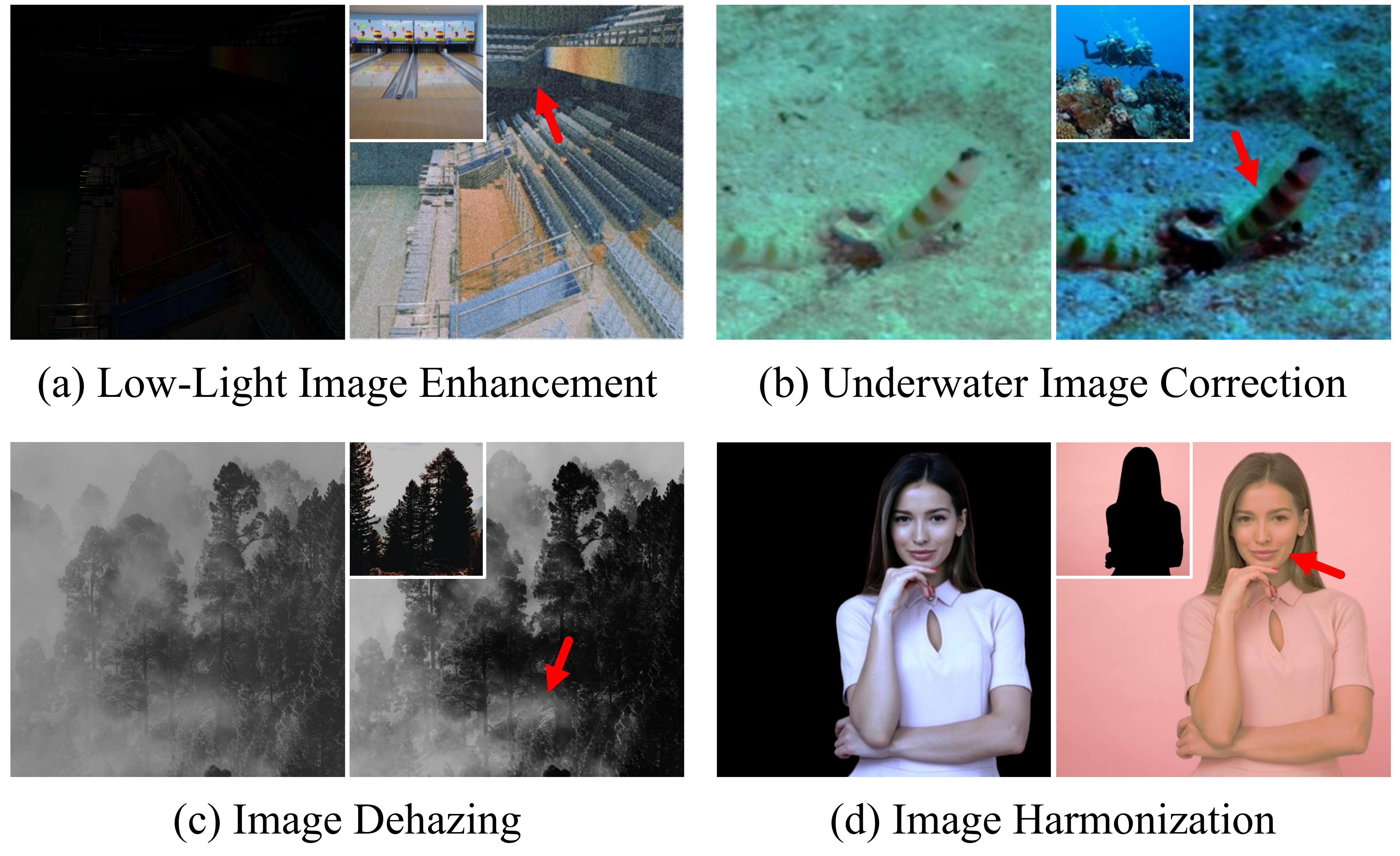}
{\begin{center}
\vspace{-0.3cm}
\caption{\textbf{Limitations of Applying Our Model to Other Tasks.} For each image pair, the left is the input image while the right is the output image. The top-left corner of the output image shows the reference image.
As indicated by red arrows: (a) heavy noise may be introduced if the input low-light image is too dark; (b) incorrect colors may be left over if the input underwater image is blurry; (c) non-uniform distributed haze in the input image may not be removed; (d) the output foreground may overfit the reference background with a solid color.
}
\label{fig:appendix_fail_other_tasks}
\end{center}
}
\vspace{-0.7cm}
\end{figure}

\;\;\;\;\;
In Fig.\,\ref{fig:othertask_supp}, we provide visual results of our Neural Preset (without fine-tuning) and other color style transfer methods on low-light image enhancement~\cite{low-light-survey}, underwater image color correction~\cite{underwater-survey}, image dehazing~\cite{dehazing-survey}, and image harmonization~\cite{harmonization-survey}. 
Neural Preset outperforms other methods by a large margin. However, since our  model is only trained in a self-supervised manner and not fine-tuned on task-specific datasets, it may fail on these tasks, 
as shown and discussed in Fig.\,\ref{fig:appendix_fail_other_tasks}. 

\begin{table*}[th]
  \begin{center}
\setlength{\tabcolsep}{16pt}
\small
\begin{tabular}{l|cccc}
\toprule
\multirow{2}{*}{Method} & \multicolumn{4}{c}{CPU Inference Time $\downarrow$} \\
\cmidrule(l){2-5} 
& \makecell[c]{FHD \\ (1920 $\times$ 1080)} & \makecell[c]{2K \\ (2560 $\times$ 1440)} & \makecell[c]{4K \\ (3840 $\times$ 2160)} & \makecell[c]{8K \\ (7680 $\times$ 4320)}   \\
\midrule
PhotoWCT~\cite{PhotoWCT} & 14.591\,s & 25.686\,s & OOM & OOM \\
WCT$^{2}$~\cite{WCT2} & 24.204\,s & 42.669\,s & OOM & OOM  \\
PhotoNAS~\cite{an2019ultrafast} & 14.227\,s & 24.826\,s & OOM & OOM  \\
Deep Preset~\cite{DeepPreset} & 14.354\,s & 25.173\,s & 58.030\,s & OOM \\
PhotoWCT$^{2}$~\cite{PhotoWCT2} & \;\;3.111\,s & \;\;4.588\,s & OOM & OOM  \\
Ours & \textbf{\;\;0.215\,s} & \textbf{\;\;0.346\,s} & \textbf{\;\;0.686\,s} & \textbf{2.290\,s} \\
\bottomrule
\end{tabular}
    \caption{\textbf{Comparison on CPU Inference Time.} 
    Evaluations are conducted on an Intel i9-11900KF CPU with 32GB PC memory. All models are in Float32 precision. 
    ``OOM'' means having the out-of-memory issue.}\label{tab:cpu_speed}
\vspace{-0.4cm}
\end{center}
\end{table*}

\vspace{-0.3cm}
\subsubsection{Comparison on CPU Inference Time}\label{appendix:compare_cpu_speed}
\vspace{-0.2cm}
\;\;\;\;\;Table\,\ref{tab:cpu_speed} shows that Neural Preset is much faster on CPU. Remarkably, it takes only 0.686 seconds to process a 4K image, but recent state-of-the-art methods either have the out-of-memory issue or take about 1 minute for processing.

\subsection{DNCM for Image Harmonization/Enhancement}\label{appendix:ncm_application}

Here we describe how to train DNCM with pairwise data for image harmonization and image color enhancement.

\medskip
\noindent\textbf{DNCM for Image Harmonization.}\quad
Extracting the foreground from one image and compositing it onto a background image is a common operation in image editing. 
In order to make the composite image more realistic, the image harmonization task is introduced to remove the inconsistent appearances between the foreground and the background. 
Recently, many image harmonization methods~\cite{DIH,S2AM,DoveNet,BargainNet,RAIN,IntrinsicIH,cvpr2022Cong,TransformerIH,Harmonizer} based on deep learning have been proposed with notable successes.

Since image harmonization can be regarded as a color mapping process from the background to the foreground inside an image, we attempt to solve it using the proposed DNCM. 
We adapt DNCM to image harmonization with two modifications. 
First, we downsample the composite image $\mathbf{I}$ and the foreground mask $\mathbf{M}$ to obtain thumbnails $\mathbf{\tilde{I}}$ and $\mathbf{\tilde{M}}$, which are concatenated as the input of the encoder $E$. 
Second, we only use DNCM to alter the color of foreground pixels (marked by $\mathbf{M}$) in $\mathbf{I}$, \ie, all background pixels in $\mathbf{I}$ are not changed.

We follow existing works to conduct experiments on the iHarmony4~\cite{DoveNet} benchmark. We evaluate the image harmonization performance by MSE and PSNR. 
The encoder $E$ is set to EfficientNet-B0~\cite{EfficientNet}, and the DNCM hyper-parameter $k$ is set to 16.
With the training loss from Tsai \etal~\cite{DIH}, our model is optimized by the Adam~\cite{Adam} optimizer for 50 epochs. We set the learning rate to $3e^{-4}$ (with a batch size of 16) and multiply it by 0.1 after every 20 epochs. 
Table\,\ref{tab:harmonization} compares our model with state-of-the-art image harmonization methods. 
Without specific modules/constraints designed for the image harmonization task, our model achieves top-level performance in terms of MSE and PSNR.
Notably, our model outperforms other methods in terms of inference time and memory footprint.

\medskip
\noindent\textbf{DNCM for Image Color Enhancement.}\quad
Image color enhancement aims to improve the visual quality of images captured in different scenes, such as underexposed or overexposed scenes. Recently, deep learning based methods~\cite{Wang_2019_CVPR,Huang2018RangeSG,hdrnet,zeng2020lut,Wang2021RealtimeIE} have dominated this field. Since we can formulate image color enhancement as a many-to-one color mapping from diverse degraded domains (\eg, underexposed and overexposed domains) to an enhanced domain, we experiment with applying DNCM to this task.

We set the DNCM hyper-parameter $k$ to 24. We train DNCM for 200 epochs using the Adam~\cite{Adam} optimizer (with a learning rate of $3e^{-4}$ and a batch size of 1). Our training loss is adopted from Zeng \etal~\cite{zeng2020lut}.
We follow Wang \etal~\cite{Wang2021RealtimeIE} to use PSNR, SSIM, and LPIPS as performance metrics. The results on the MIT-Adobe FiveK benchmark~\cite{BychkovskyPCD11} (Table\,\ref{tab:enhancement}) demonstrate that our model performs on par with the state-of-the-art methods. The inference speed of our model is also comparable to the methods designed to run in real time~\cite{zeng2020lut,Wang2021RealtimeIE}.

\subsection{On-Device Deployment of Neural Preset}\label{appendix:distributed}

On-device~\cite{on_device1} (\eg, mobile or browser) applications expect a small computational overhead in the client to support real-time UI responses and a small amount of data exchanges between the client/server to save network bandwidth. However, existing color style transfer models are not suitable for such applications: deploying them in the client requires too much memory and is computationally expensive, while deploying them in the server will significantly increase the network bandwidth since high-resolution images must be transmitted over the internet.
Instead, our Neural Preset supports distributed deployment to alleviate this problem: we can deploy the encoder $E$ in the server and  \textit{nDNCM}/\textit{sDNCM} in the client. In this way, the client-side calculation can be fast. Besides, only thumbnails and the DNCM parameters are transmitted over the internet.


\begin{table}[t]
  \begin{center}
\setlength{\tabcolsep}{6pt}
\small
\begin{tabular}{l|cc|c}
\toprule
\multirow{2}{*}{Method} & \multicolumn{2}{c}{Performance} & GPU Inference \\
\cmidrule(lr){2-3}  \cmidrule(lr){4-4}
& MSE $\downarrow$  & PSNR $\uparrow$ & Time $\downarrow$ / Memory $\downarrow$  \\
\midrule
S$^2$AM~\cite{S2AM} & 59.67 & 34.35 & 0.148\,s / \;\;6.3\,GB \\
DoveNet~\cite{DoveNet} & 52.36 & 34.75 & 0.072\,s /  \;\;6.5\,GB \\
BargainNet~\cite{BargainNet} & 37.82 & 35.88 & 0.086\,s / \;\;3.7\,GB \\
IntrinsicIH~\cite{IntrinsicIH} & 38.71 & 35.90 & 0.833\,s / 16.5\,GB \\
IHT~\cite{TransformerIH} & 37.07 & 36.71 & 0.196\,s / 18.5\,GB \\
Harmonizer~\cite{Harmonizer} & 24.26 & 37.84 & 0.017\,s / \;\;2.3\,GB\\
CDTNet~\cite{cvpr2022Cong} & \textbf{23.75} & \textbf{38.23} & 0.023\,s / \;\;8.1\,GB\\
\midrule
DNCM (Ours) & 24.31 & 37.97 &  \textbf{0.006\,s} / \;\;\textbf{1.1\,GB} \\
\bottomrule
\end{tabular}
\caption{\textbf{Image Harmonization Results on iHarmony4.} 
    The performance metrics (MSE and PSNR) are computed at $256 \times 256$ resolution, while the inference time and memory footprint are measured at Full HD resolution on a Nvidia RTX3090 GPU.
    }\label{tab:harmonization}
\vspace{-0.2cm}
\end{center}
\end{table}

\begin{table}[t]
  \begin{center}
\setlength{\tabcolsep}{7pt}
\small
\begin{tabular}{l|ccc}
\toprule
Method & PSNR $\uparrow$ &  SSIM $\uparrow$ & LPIPS $\downarrow$ \\
\midrule
UPE~\cite{Wang_2019_CVPR} & 20.03 & 0.7841 & 0.2523 \\
RSGUNet~\cite{Huang2018RangeSG} & 21.37 & 0.7998 & 0.1861 \\
HDRNet~\cite{hdrnet} & 22.15 & 0.8403 & 0.1823 \\
Adaptive 3D LUT~\cite{zeng2020lut} & 22.27 & 0.8368 & 0.1832 \\
Learnable 3D LUT~\cite{Wang2021RealtimeIE} & \textbf{23.17} & 0.8636 & 0.1451 \\
\midrule
DNCM (Ours) & 23.12 & \textbf{0.8697} & \textbf{0.1439} \\
\bottomrule
\end{tabular}
    \caption{\textbf{Image Color Enhancement Results on FiveK.} All performance metrics are calculated at the Full resolution, \ie, the original resolution of the test samples.
    }\label{tab:enhancement}
\vspace{-0.3cm}
\end{center}
\end{table}

\begin{figure}[t]
\centering
\includegraphics[width=0.9\linewidth]{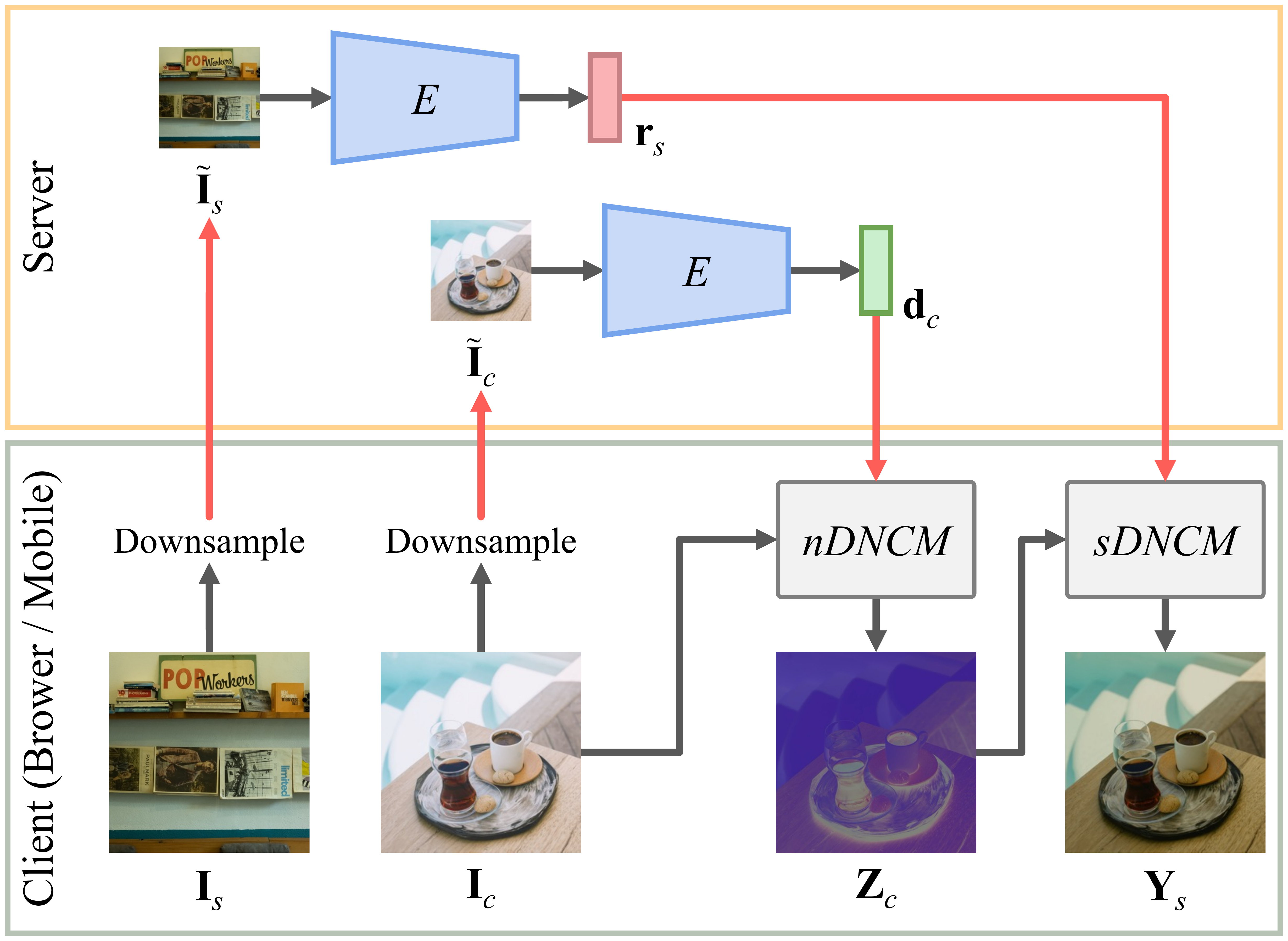}
{\begin{center}
\vspace{-0.3cm}
\caption{\textbf{On-Device Deployment of Neural Preset.} 
Black arrows represent server or client processing flow, while red arrows represent network transmission between server and client.
}
\label{fig:on_device_deployment}
\end{center}
}
\vspace{-0.7cm}
\end{figure}

As illustrated in Fig.\,\ref{fig:on_device_deployment}, 
when transferring color style from a style image $\mathbf{I}_s$ to an input image $\mathbf{I}_c$, the client first downsamples the two images and transmits their thumbnails to the server. Note that the file size of a thumbnail with a resolution of $256 \times 256$ is only about 30KB, but the file size of a 4K resolution image can be up to 20MB. Then, the server calculates the color style parameters $\mathbf{d}_c$/$\mathbf{r}_s$ (the data size is about 3KB) from the uploaded thumbnails and transmits $\mathbf{d}_c$/$\mathbf{r}_s$ back to the client. 
Finally, the client performs \textit{nDNCM}/\textit{sDNCM} on the high-resolution $\mathbf{I}_c$ to complete color style transfer, which is fast and require only a small memory footprint. 

\section*{Acknowledgments}
Most of the images we displayed in this paper are from the \textit{pexels.com} and \textit{flickr.com} websites, which are under the Creative Commons license. The rest of the images we displayed are from publicly available datasets \cite{dataset_RESIDE,dataset_REVIDE,openimage,dataset_LOL,dataset_EUVP,dataset_UIEB,dataset_UIMAGENET,DoveNet,BychkovskyPCD11, MSCOCO}. 
We thank the artists and photographers for sharing their amazing works online, and we thank the researchers who constructed the datasets.
Besides, we thank Mia Guo for her efforts in taking and retouching the photo of Wukang Mansion, Shanghai, China  ({\it aka} I.S.S Normandie Apartment) used in Fig.\,\ref{fig:filterlutcomp}.

We thank the project contributors (Weiwei Chen, Jing Li, and Xiaojun Zheng) for their help in developing demos.
We also appreciate all user study participants and anonymous peer reviewers.


{\small
\bibliographystyle{ieee_fullname}
\bibliography{egbib}
}

\end{document}